\documentclass[conference]{IEEEtran}
\IEEEoverridecommandlockouts
\usepackage{fancyhdr}
\usepackage[T1]{fontenc}
\usepackage{mathtools}

\newcommand{\bvf}{{\mathbf{f}}}
\newcommand{\bvh}{{\mathbf{h}}}

\newcommand{\bR}{{\mathbf{R}}}

\newcommand{\rv}[1]{\boldsymbol{#1}}

\usepackage{hyperref}
\usepackage{xcolor}
\hypersetup{
    colorlinks,
    linkcolor={black},
    linkcolor={black},
    citecolor={black},
    urlcolor={blue!80!black}
}

\usepackage{graphicx}
\usepackage{amsmath}
\usepackage{multirow}
\usepackage{amssymb}
\usepackage{wrapfig}
\usepackage{array}
\usepackage{soul}
\usepackage{xcolor}
\usepackage{verbatim}
\usepackage{subcaption}
\usepackage{tikz}
\usetikzlibrary{spy}
\usepackage{algorithm}
\usepackage{textcomp}
\usepackage[export]{adjustbox}
\usepackage{siunitx}
\usepackage{array}
\usepackage{commath}
\usepackage{url}
\usepackage{paralist}
\usepackage{booktabs} % For formal tables
\usepackage{algpseudocode}
\usepackage{gensymb}
\usepackage{enumitem}
\usepackage{cancel}
\usepackage{todonotes}
\usepackage{pdfpages}
\usepackage{animate}
\usepackage{cite}
\usepackage[rightcaption]{sidecap} % Load in preamble

\setlist{nolistsep}
\graphicspath{ {./figs/} }
\usepackage[switch]{lineno}

\def\BibTeX{{\rm B\kern-.05em{\sc i\kern-.025em b}\kern-.08em
    T\kern-.1667em\lower.7ex\hbox{E}\kern-.125emX}}

\begin{document}

\title{\LARGE Global Attention with Linear Complexity for Exascale Generative Data Assimilation in Earth System Prediction\thanks{This manuscript has been authored by UT-Battelle, LLC, under contract DE-AC05-00OR22725 with the US Department of Energy (DOE). The US government retains and the publisher, by accepting the article for publication, acknowledges that the US government retains a nonexclusive, paid-up, irrevocable, worldwide license to publish or reproduce the published form of this manuscript, or allow others to do so, for US government purposes. DOE will provide public access to these results of federally sponsored research in accordance with the DOE Public Access Plan.}}

\author{
\IEEEauthorblockN{Xiao Wang\IEEEauthorrefmark{1},
Zezhong Zhang\IEEEauthorrefmark{2},
Isaac Lyngaas\IEEEauthorrefmark{1},
Hong-Jun Yoon\IEEEauthorrefmark{1},
Jong-Youl Choi\IEEEauthorrefmark{1},
Siming Liang\IEEEauthorrefmark{1},
Janet Wang\IEEEauthorrefmark{1}\\
Hristo G.~Chipilski\IEEEauthorrefmark{5},
Ashwin M. Aji\IEEEauthorrefmark{3},
Feng Bao\IEEEauthorrefmark{5},
Peter Jan van Leeuwen\IEEEauthorrefmark{6},
Dan Lu\IEEEauthorrefmark{1},
Guannan Zhang\IEEEauthorrefmark{7}
}
\vspace{0.1cm}
\IEEEauthorblockA{\IEEEauthorrefmark{1}Oak Ridge National Laboratory, Oak Ridge, TN, USA,
\{wangx2,lyngaasir, yoonh, liangs1,  choij, wangs3, lud1\}@ornl.gov}
\IEEEauthorblockA{\IEEEauthorrefmark{2}Auburn University, Auburn, AL, USA, zez0002@auburn.edu}
\IEEEauthorblockA{\IEEEauthorrefmark{5} Florida State University, Tallahassee, FL, USA, 
\{fbao, hchipilski\}@fsu.edu}
\IEEEauthorblockA{\IEEEauthorrefmark{3}AMD Research and Advanced Development, Santa Clara, CA, USA, ashwin.aji@amd.com}
\IEEEauthorblockA{\IEEEauthorrefmark{6} Colorado State University, Fort Collins, CO, USA, 
peter.vanLeeuwen@colostate.edu}
\IEEEauthorblockA{\IEEEauthorrefmark{7} Corresponding author, Oak Ridge National Laboratory, Oak Ridge, TN, USA, zhangg@ornl.gov}
}

\maketitle

%%%%%%%%% ABSTRACT
\begin{abstract} %150 WORDS
Accurate Earth system prediction requires state inference from incomplete observations, but conventional two-stage data assimilation (DA) is computationally prohibitive because repeated PDE-based ensemble forecasts, observation updates, and intermediate data movement limit ensemble size at high resolution. We introduce STORM, a one-stage generative AI framework that reformulates DA as diffusion-based Bayesian posterior sampling, replacing online PDE ensemble forecasts with scalable AI inference. It further combines a spatiotemporal transformer with a global-attention algorithm that reduces complexity from quadratic to linear through scalable gradient propagation, enabling high-resolution, long-context Earth modeling.  STORM scales to 74,400 GPUs on Frontier with 96--99\% strong-scaling efficiency and up to 6 ExaFLOPs sustained BF16 throughput, while enabling 32,768-member ensembles for uncertainty quantification in 34 seconds on 4,096 GPUs. It scales to 20 billion spatiotemporal tokens and 177,000 temporal frames. Hurricane tracking and long-term climate reanalysis demonstrate improved accuracy, including benefits from longer temporal context and recovery of temperature extremes missed by forecast-only predictions.
\end{abstract}

%%%%%%%%% BODY TEXT
% \input{./justification}
\section{Problem Overview}
\label{sec:problem_overview}

Accurate Earth system prediction requires two complementary capabilities:
\emph{forward simulation}, which predicts how the Earth evolves from an
estimated initial state, and \emph{state inference}, which estimates the
actual evolving Earth system state by combining the forward simulation with incomplete, noisy, and heterogeneous observations. These capabilities form a continuous loop: state inference uses forward simulation to constrain possible states, while inferred state provides a better initial condition for
subsequent forward simulation. Thus, forward simulation alone cannot provide a
complete Earth system modeling capability; without scalable state inference,
simulations may evolve according to the model but diverge from the observed
Earth system.

\emph{Data assimilation} (DA) provides this state inference capability by combining a forward simulation with observational information to infer the Earth
system state through Bayesian inference~\cite{houtekamer2016review}. DA is essential for weather prediction because the streaming observations must continuously correct the estimated weather state, particularly for extreme events such as hurricanes. DA is equally important to climate science through reanalysis: observationally constrained, spatially complete reconstructions of the historical Earth system that support the analysis of climate variability and trends, extreme events, prediction initialization, and climate-model evaluation~\cite{era5, buizza2018advancing}.

\textbf{Forward simulation has reached exascale; DA has not.}
Recent advances in supercomputing have dramatically advanced forward
simulation through high-fidelity Earth System Models
\cite{E3SMMMF2023,ClimateExascale2025,GCM2024Winner,taylor2023simple,klocke2025computing}
and AI-based forecasting models
\cite{GraphCast2023,PanguWeather2023,FourCastNet2022,Gencast}. These
approaches increasingly enable global Earth simulation
at high spatial resolution. 
In contrast, DA has not yet achieved comparable exascale scalability, particularly for high-resolution, large-ensemble uncertainty quantification (UQ), and low-latency workflows. Specifically, the following two scaling barriers prevent DA from scaling effectively:

{\bf Barrier 1: Two-stage DA workflow}. Bayesian DA combines two fundamentally different operations: (1) partial differential equation (PDE)-based forward simulation, which forecasts the evolution of the Earth system, and (2) observational inference, which uses observations to update the predicted state from (1). Conventional DA performs the above two operations sequentially in a two-stage forecast-update cycle~\cite{houtekamer2016review}. This coupled workflow requires repeatedly running expensive PDE forward simulations across large ensembles and then passing the resulting high-dimensional forecast states to the observational update. The resulting computational and data-movement costs grow rapidly with spatial resolution and ensemble size, limiting high-resolution, large-ensemble, and low-latency DA even on exascale systems.

{\bf Barrier 2: Scaling global spatiotemporal AI}. Generative AI provides an opportunity to replace the expensive PDE-based ensemble forward simulation with AI inference, which is orders of magnitude cheaper. For high-dimensional Earth data, \emph{Vision Transformers} (ViTs) provide a flexible architecture for representing spatial and temporal fields and heterogeneous Earth 
states~\cite{nguyen2023climax,wang2024orbitoakridgebase,2025orbit2,prithvi}. However, replacing PDE with generative AI shifts the computational
bottleneck to the scalability of the AI model itself. With $N$ spatial tokens per frame and $K$ temporal frames, conventional global attention over the complete spatiotemporal sequence has {\bf \em quadratic complexity} with respect to $NK$. Increasing either spatial resolution or temporal context therefore rapidly increases computation and memory, forcing existing ViT approaches to trade among spatial resolution, temporal context, and global interactions.

{\bf Exascale generative AI for DA.} We address both scaling barriers through STORM (\emph{{S}patiotemporal {T}ransformer for {O}bservation-informed {R}econstruction and {M}odeling}), a unified exascale generative DA framework that integrates one-stage diffusion-based posterior sampling with a linear-complexity global attention spatiotemporal ViT.

STORM addresses Barrier 1 through STORM-OneStage, \emph{a one-stage
generative DA formulation}. Instead of first generating a large forecast
ensemble and then transferring it to a separate observational update, STORM directly samples posterior states conditioned jointly on historical context and observations. The learned generative model replaces the PDE-based forecast ensemble generation,
substantially reducing the cost of DA, while direct posterior generation
eliminates storage, data movement and expensive I/O operations for the intermediate forecast ensemble. Moreover,  STORM conditions
inference on many historical states, enabling long temporal context that
improves long-horizon climate reanalysis (e.g., STORM enables
32,768-member posterior ensembles in 34 seconds on 4,096 GPUs.)

STORM addresses Barrier 2 through STORM-DiT, \emph{a scalable
spatiotemporal diffusion-transformer architecture}, and
STORM-GPA, \emph{a global-attention algorithm based on scalable gradient
propagation}. STORM-DiT reduces spatiotemporal attention complexity from $\mathcal{O}(N^2K^2)$ to $\mathcal{O}(N^2)$, where $N$ is the number of spatial tokens per temporal frame and $K$ is the number of frames, while STORM-GPA further reduces global-attention complexity to $\mathcal{O}(N)$ while preserving global
interactions. Together, they enable 20 billion spatiotemporal tokens and
177k temporal frames.

In short, three contributions are summarized below:
(1) {\bf unified one-stage generative DA}, enabling 32,768-member UQ in
34 seconds on 4,096 GPUs;
(2) {\bf linear-complexity global spatiotemporal modeling} through
STORM-DiT and STORM-GPA, enabling 20 billion tokens and 177k temporal frames; and
(3) {\bf an exascale scientific capability}, scaling STORM to
74,400 Frontier GPUs with 96--99\% strong-scaling efficiency and up to
6 ExaFLOPs sustained BF16 throughput.
These capabilities enable observation-constrained hurricane tracking and
climate reanalysis, including reconstruction of extreme events missed by the forecast prior. Moreover, multi-year temporal context improves climate reanalysis accuracy, linking computational scalability to new science capability.

\section{Background \& State of The Art}
\label{sec:SOA}
\subsection{Conventional Two-Stage DA and Scaling Strategies} 
As discussed in Sec.~II, Barrier 1 arises from the conventional two-stage forecast–update DA workflow. Here, we review how existing DA methods make this workflow computationally tractable and why these strategies alleviate, but do not remove, the underlying scaling barrier. 

DA estimates the evolving Earth system state by combining two complementary sources of information: a forward simulation model that forecasts possible future Earth system states from the previously estimated state, and an observation model that constrains which of these possible states are consistent with observations. For the forward simulation forecast, the Earth system state $\rv x_k$ at time $k$ is
estimated from the previous state $\rv x_{k-1}$ according to:
\begin{equation}\label{state}
\rv x_k = \bvf(\rv x_{k-1}, \rv \varepsilon^m_k) \ ,
\end{equation}
where $\bvf$ is the forward simulation model and $\rv \varepsilon^m_k$ represents model uncertainty. Repeated forward simulations define the conditional
\emph{prior distribution}
$p(\rv x_k|\rv x_{k-1})$, represented by an ensemble of estimated states using the above equation.

In the update stage, new observations $\rv y_k$ relates to the state through observation model:
\begin{equation}\label{observations}
\rv y_k = \bvh(\rv x_k) + \rv \varepsilon^o_k \ ,
\end{equation}
where $\bvh$ is the observation operator that maps the Earth system state to
the observation space and $\rv \varepsilon^o_k$ represents observation error. 
This observation model defines the \emph{likelihood function}
$p(\rv y_k|\rv x_k)$.
Because the previous state is itself uncertain, the prior distribution at time $k$ is obtained by marginalizing over the distribution of $\rv x_{k-1}$. Combined with the likelihood through Bayes' rule,the posterior $p(\rv x_k | \rv y_k)$:
\begin{equation}\label{Bayesian}
p(\rv x_k | \rv y_k)
\propto
p(\rv y_k|\rv x_k)
\int
p(\rv x_k|\rv x_{k-1})
p(\rv x_{k-1})d\rv x_{k-1}.
\end{equation}
The posterior represents the ensemble of Earth system states that are both consistent with the forward simulation and the observations. This posterior becomes the starting point for the next forecast stage, and the cycle repeats as new observations arrive. This repeated two-stage structure underlies most operational DA systems~\cite{houtekamer2016review}.

{\bf State-of-the-art DA methods mitigate, rather than eliminate, Barrier 1}.
At a kilometer-scale global resolution, the full Earth system state contains
$\mathcal{O}(10^{10})$ degrees of freedom. Reliable forecast UQ requires repeated forward simulations to generate a large ensemble. However, operational ensemble systems are typically limited to tens to hundreds of members because of computational and latency constraints \cite{palmer_2018}, whereas much larger ensembles may be required to characterize complex dynamical systems, distribution tails, and rare events \cite{palmer_2018}.

Finite ensemble sizes introduce sampling error and motivate approximations such as covariance localization \cite{morzfeld2023theory, houtekamer2016review}. Practical DA methods may also rely on Gaussian or near-Gaussian assumptions that simplify posterior inference but can limit the representation of strongly non-Gaussian uncertainty \cite{bocquet2010beyond}. These techniques have been essential for making DA more computationally feasible, but they do not change the underlying forecast–update structure: a forecast ensemble must still be generated before observations are incorporated, and the resulting high-dimensional states must subsequently be processed by the update stage.

Operational latency further constrains this workflow. Rapid-refresh systems must complete forecast and assimilation cycles within strict time windows; for example, High-Resolution Rapid Refresh (HRRR) operates on an hourly updating cycle \cite{dowell2022hrrr}. Increasing spatial resolution or ensemble size therefore increases computational demand where additional latency is least affordable. Thus, conventional approximations, numerical improvements, and parallelization reduce particular costs, but Barrier 1 remains: a large forward simulation ensemble must be repeatedly generated before observational update. Removing this barrier requires changing the mathematical and computational paradigm of DA rather than only optimizing its individual components.

\subsection{AI-Based Forward Simulation and Generative DA}
Recent AI models substantially reduce one major component of Barrier 1: the cost of forward simulation. GraphCast~\cite{GraphCast2023}, Pangu Weather~\cite{PanguWeather2023}, FourCastNet~\cite{FourCastNet2022}, and GenCast~\cite{Gencast} have demonstrated fast global weather forecasting, with inference achievable in seconds on a small number of GPUs. These advances show that AI can provide an efficient alternative to repeated PDE-based forward simulation.

{\bf However, accelerating or replacing the PDE solver for forward simulation alone does not eliminate Barrier 1.} Forward simulations propagate a previously estimated state into the future but do not incorporate new observations to infer the evolving Earth system state. Replacing the PDE solver with an AI forward model can therefore substantially reduce the cost of generating the prior ensemble, but the subsequent observational inference remains. More recently, diffusion-based generative models have been explored for the forecast step of DA~\cite{aida1,aida2,aida3s},
including our previous work on score-based filtering
\cite{Bao2024ScoreFilter,Bao2024EnsembleScore,Bao2025DiffusionFiltering}. These approaches improve computational efficiency and provide flexible representations of nonlinear and non-Gaussian uncertainty. Nevertheless, they generally retain a sequential two-stage workflow in which a prior distribution is first generated and observations are subsequently incorporated through an update stage.

Consequently, these methods improve important components of conventional DA but do not fundamentally remove the intermediate forecast distribution connecting prediction and observational inference. At large ensemble sizes and high spatial resolution, the forecast states must still be generated, represented, and processed before the observational update can be completed.
To eliminate Barrier 1 rather than just reduce its cost, posterior inference should avoid explicitly generating a large intermediate forward simulation ensemble and instead incorporate dynamical history and observations within a unified inference process. This is the motivation for the STORM-OneStage framework introduced in Sec.~\ref{subsec:storm-OneStage}.

\subsection{Transformer Scaling and the Global-Context Trade-off}
{\bf Addressing Barrier 1 with generative AI exposes the challenges associated with Barrier 2, i.e., scaling global spatiotemporal AI to simultaneously support high spatial resolution and long temporal context.} High-resolution Earth system modeling involves $\mathcal{O}(10^{10})$ spatial degrees of freedom, while long temporal context may require jointly learning tens to thousands of historical frames. Such spatiotemporal context is important for representing dependencies spanning multiple time scales in weather and climate analysis, such as hurricane tracking or long-horizon climate event prediction. 

ViT have emerged as a leading architecture for large-scale Earth system foundation models because their token-based representation provides a flexible mechanism for modeling heterogeneous spatial, temporal, and multivariable fields \cite{nguyen2023climax,wang2024orbitoakridgebase,2025orbit2,prithvi}. ViTs represent spatiotemporal fields as sequences of tokens, where each token corresponds to a localized spatial patch at a given time frame. With $K$ temporal frames and $N$ spatial tokens per temporal frame, the total sequence length is $K\times N$. The attention mechanism of ViTs then computes global pairwise interactions
among all tokens, resulting in {\bf \em quadratic} computational and memory
complexity with respect to the spatiotemporal sequence length. Consequently, simultaneously increasing global spatial resolution and temporal context rapidly becomes computationally prohibitive. For example, scaling to
$\mathcal{O}(10^9)$ tokens requires $\mathcal{O}(10^{18})$ pairwise token interactions and exabyte-scale attention memory, far exceeding the capabilities of current exascale systems.

\begin{figure}[h!]
\centering
\includegraphics[width=0.95\linewidth]{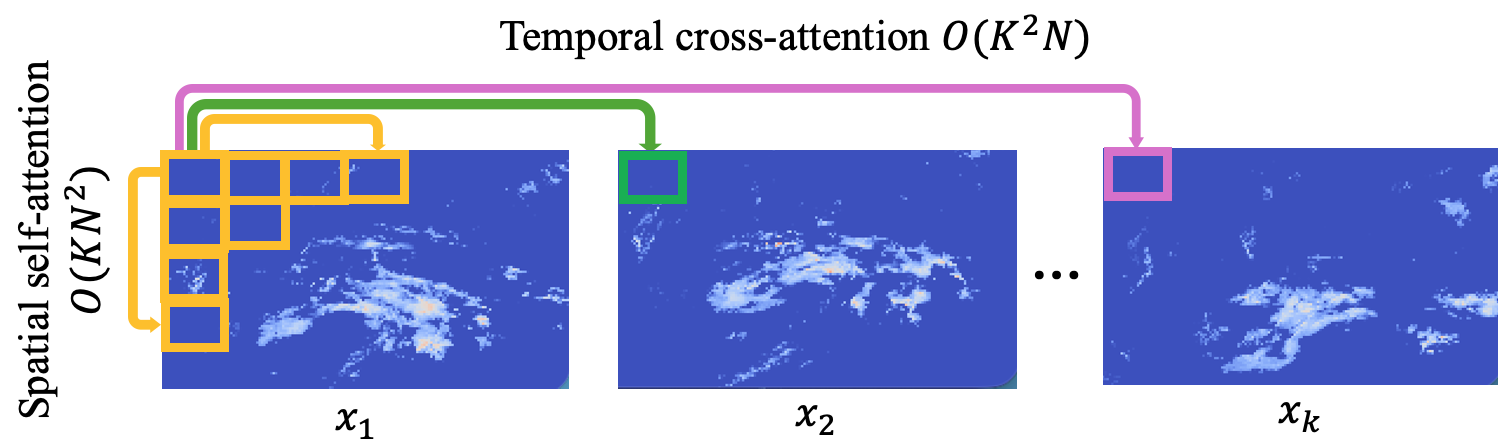}
\caption{\em An illustration of the TimeSFormer operations with $\mathcal{O}(K^2N+KN^2)$ complexity.}
\label{fig:TimeSFormer}
% \vspace{-0.1cm}
\end{figure}

{\bf Existing transformer approaches address different aspects of Barrier 2, but introduce significant trade-offs.} One strategy is to restrict attention to local spatial regions. Methods such as Swin Transformer~\cite{liu2022swinvit2}, MaxViT~\cite{maxvit} and ORBIT-2~\cite{2025orbit2} restrict attention to local spatial windows. ORBIT-2, in particular, scaled to 4 billion spatial tokens but by limiting interactions to local spatial regions, and does not jointly model long temporal context. Localization therefore enables large-scale computation, but direct global interactions are sacrificed. A second strategy preserves global interactions while factorizing spatial and temporal attention. TimeSformer~\cite{bertasius2021}, a state-of-the-art ViT, factorizes attention into temporal and spatial components with per-layer complexity $\mathcal{O}(K^2N + KN^2)$. As illustrated in Fig.~\ref{fig:TimeSFormer}, this corresponds to computing temporal attention at each spatial location (different color squares in the figure) followed by spatial attention at each time frame (same color squares). Despite this factorization, the quadratic dependence on both space and time makes large deployment infeasible, with TimeSformer limited to $\sim$3 million tokens, far below the required billion-token regime.
\begin{table}[h!]
    \footnotesize
    \centering
    \begin{tabular}{lccc}
    \toprule
    \textbf{Method} & \textbf{Complexity} & \textbf{Max Tokens} \\
    \midrule
    ViT (global) & $\mathcal{O}(K^2N^2)$ & $\sim 10^6$  \\
    TimeSformer (global) & $\mathcal{O}(N^2K+K^2N)$ & $\sim 3 \times 10^6$  \\
    ORBIT-2 (local, spatial only) & $\mathcal{O}(N)$ & $\sim 4 \times 10^9$  \\
    \bottomrule
    \end{tabular}
    \caption{\em Comparison of complexity and scalability across spatiotemporal AI methods. Existing localized methods achieve linear complexity, but lose global attention.}
    \label{Table:complexity-comparison}
    % \vspace{-0.5cm}
\end{table}

Table~\ref{Table:complexity-comparison} summarizes this trade-off. 
Existing approaches either achieve scalability by restricting interactions to local regions or preserve global interactions while retaining quadratic scaling. Thus, incremental modifications to conventional attention do not overcome {Barrier 2}, namely no existing approach simultaneously provides global interactions, high spatial resolution, long temporal context, and scalable computational complexity.

\section{Innovation Realized}
\label{sec:innovation}
\begin{figure*}[t]
    \centering    
\includegraphics[width=0.8\linewidth]{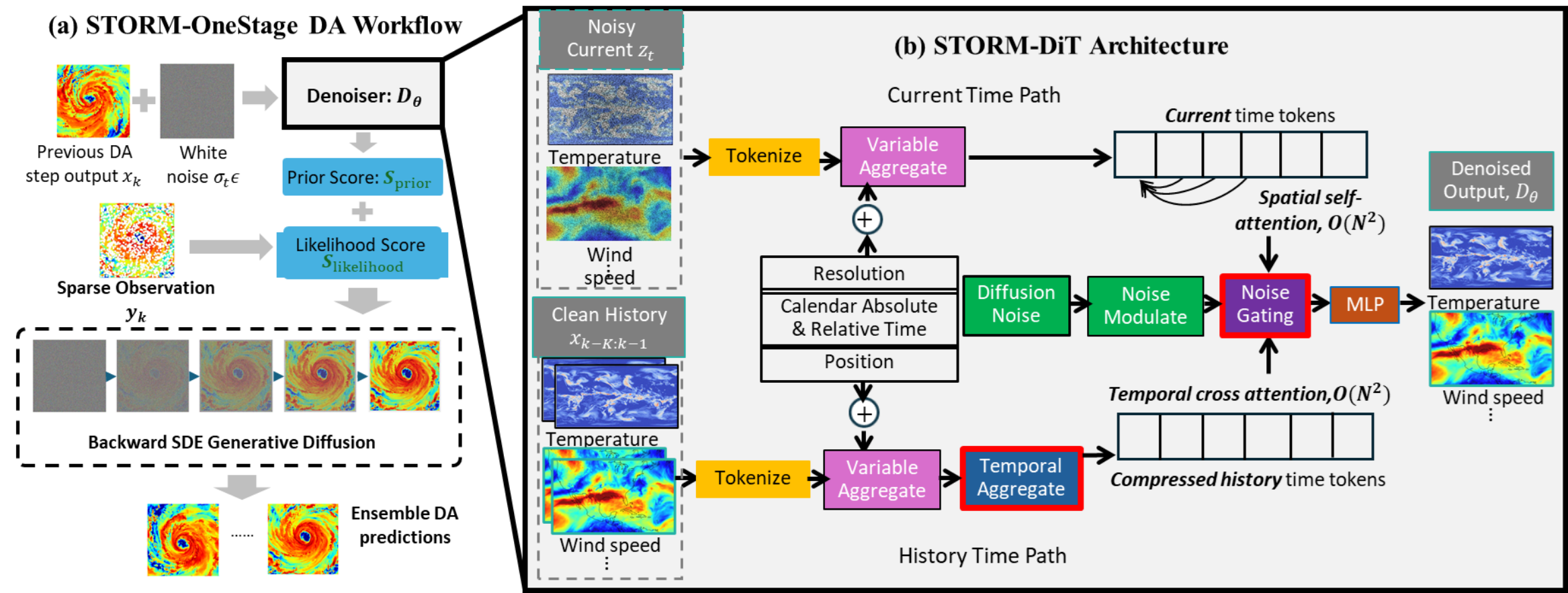} 
\caption{\em (a) STORM-OneStage generative DA workflow directly samples posterior Earth system states from historical context and observations. (b) STORM-DiT architecture. Historical states are compressed into a global temporal representation, while the current state remains at full resolution. Noise-gated spatial and temporal attention decouple space–time interactions, reducing complexity from $\mathcal{O}(K^2N + KN^2)$ to $\mathcal{O}(N^2)$ while preserving global correlations.}
    \label{fig:workflow-architecture}
    \vspace{-0.3cm}
\end{figure*}

\subsection{STORM-OneStage: Unified One-Stage Generative DA}
\label{subsec:storm-OneStage}
The conventional two-stage DA requires the repeating two-stage forecast-update cycles at every
assimilation time, generating an intermediate forecast ensemble and passring it to a separate observation update stage.
STORM fundamentally changes this computational paradigm to a unified one-stage DA by inferring the current Earth system state from observations and long-context historical states simultaneously using a conditional generative AI model.

Specifically, STORM samples the current state $\rv x_k$ conditioned on
observations $\rv y_k$ and $K$ historical states
$\rv x_{k-K:k-1}$ from the posterior:
\begin{equation}\label{eq:post2}
p(\rv x_k | \rv y_k,\rv x_{k-K:k-1})
\propto
p(\rv y_k | \rv x_k)
p(\rv x_k | \rv x_{k-K:k-1}) \ ,
\end{equation}
where $p(\rv y_k | \rv x_k)$ denotes the observational 
likelihood function, $p(\rv x_k | \rv x_{k-K:k-1})$ is the conditional prior distribution given the
$K$ frames of historical states, and it is represented by a learned generative diffusion model.

This formulation provides three key benefits. First, the scalable generative sampling replaces repeated and expensive online PDE-based forecast computation, substantially reducing the cost of posterior sampling. Second, direct posterior generation eliminates the intermediate forecast ensemble and its associated data storage, movement and I/O operations between the two stages, enabling much larger ensembles for reliable UQ. Third, unlike many conventional DA workflows that advances the system sequentially from one historical state at each assimilation step, STORM can condition inference jointly on many historical states. This long temporal context improves long-term climate reanalysis accuracy, which will be demonstrated later in experiment Figs.~\ref{fig:world}--\ref{fig:climate_event}.

\vspace{0.1cm}
\textbf{Diffusion-based posterior sampling.}
STORM represents the conditional prior $p(\rv x_k|\rv x_{k-K:k-1})$ with a diffusion model~\cite{Karras2022edm}. The diffusion model is trained to learn how to reverse a predefined noise-adding process, so that it can generate samples from the learned conditional prior. During posterior inference, the learned prior is combined with the observation likelihood to guide the same denoising process toward states consistent with the observations. The workflow is illustrated in Fig.~\ref{fig:workflow-architecture}(a).

When training the diffusion model, a forward stochastic differential equation (SDE)
progressively adds Gaussian noise to the state $\rv x_k$ from
diffusion steps $t=0$ to $t=T$, i.e., 
\begin{equation}\label{forward:SDE}
d\rv z_t =
\sqrt{2\sigma'_t\sigma_t}\,d\rv W_t .
\end{equation}
The resulting noisy state is
$
\rv z_t=\rv x_k+\sigma_t\boldsymbol{\epsilon}, \quad
\boldsymbol{\epsilon}\sim\mathcal{N}(\mathbf{0},\mathbf{I}),
$
where $\sigma_t$ controls the noise level and increases with $t$. Thus,
$\rv z_0=\rv x_k$ and $\rv z_T$ approaches the standard Gaussian noise. STORM is
trained to learn the reverse denoising dynamics by
learning the score function of the conditional prior.

During inference, STORM starts from a standard Gaussian noise and uses the learned reverse denoising dynamics to generate a state from the conditional prior. To obtain a posterior sample rather than a prior sample, the observation likelihood is
incorporated into the reverse process. Specifically, the reverse SDE progressively denoises $\rv z_t$ from $t=T$ to $t=0$:
\begin{equation}\label{reverse:SDE}
d\rv z_t = -2 \sigma'_t \sigma_t \, \rv S_{\text{posterior}}(\rv z_t,t|\rv y_k)  dt + \sqrt{2 \sigma'_t \sigma_t}  d \rv W_t,
\end{equation}
where $\rv S_{\mathrm{posterior}}$ is the gradient of the log posterior (i.e., the posterior score), and
guides the denoising trajectory toward states that are both consistent with the learned prior and the observations. The resulting
trajectory generates a sample from the desired posterior distribution in Eq.~\eqref{eq:post2}.

\vspace{0.1cm}
\textbf{Posterior score decomposition.} The reverse SDE in Eq.~\eqref{reverse:SDE} requires the posterior score,
which combines the learned conditional prior with the observation
data. Rather than learning a separate posterior score for each
observation, we decompose it into a prior score $\rv S_{\mathrm{prior}}$ and a likelihood score $\rv S_{\mathrm{likelihood}}$:
\begin{equation}\label{score_decomp}
\rv S_{\mathrm{posterior}}
=
\rv S_{\mathrm{prior}}
+
\rv S_{\mathrm{likelihood}} \ ,
\end{equation}
where $\rv S_{\mathrm{prior}}$ is the gradient of the log conditional prior
and captures the physical structure of the Earth system. STORM learns this
prior score with its scalable STORM-DiT architecture
(see Sec.~\ref{sec:STORM-architecture}), providing the learned replacement
for the conventional PDE-based forecast model. The likelihood score
$\rv S_{\mathrm{likelihood}}$ represents the gradient of the log
observation likelihood and incorporates the available observations.

\textbf{Efficient likelihood score approximation.}
The likelihood score incorporates the observations but is generally
unavailable at the noisy intermediate states $\rv z_t$ during diffusion.
We therefore use Diffusion Posterior Sampling (DPS) to approximate the
likelihood score from the denoised estimate
$\hat{\rv z}_0(\rv z_t)$. Under Gaussian observation noise,
\begin{equation}\label{dps_final}
\hspace{-0.2cm}\rv S_{\mathrm{likelihood}}(\rv y_k|\rv z_t,t)
\approx
-\zeta_t
\nabla_{\rv z_t}
\left\|
\rv y_k-\bvh(\hat{\rv z}_0(\rv z_t))
\right\|^2_{\bR_k^{-1}},
\end{equation}
where $\bvh$ maps the estimated Earth system state to the observation and $\zeta_t$ controls the strength of the observational correction. $\bR_k$ is covariance of observation errors.
Thus, each reverse SDE step combines the learned prior
with the available observations to progressively generate a posterior
sample.

\subsection{Scalable Spatiotemporal STORM-DiT Architecture}\label{sec:STORM-architecture}
The one-stage DA formulation requires the score of the conditional prior $p(\rv x_k | \rv x_{k-K:k-1})$, which replaces the conventional PDE-based forecast model in posterior sampling. A key observation is that this prior score can be estimated through a denoising function. Specifically, Tweedie's identity gives:
\begin{align}
\hspace{-0.1cm}
\rv S_{\text{prior}}(\rv z_t, \sigma_t | \rv x_{k-K:k-1}) 
&= \frac{\mathbb{E}[\rv z_0 | \rv z_t, \rv x_{k-K:k-1}] - \rv z_t}{\sigma_t^2} \\
&\approx \frac{D_\theta(\rv z_t, \sigma_t ; \rv x_{k-K:k-1}) - \rv z_t}{\sigma_t^2}, \label{eqn:denoise}
\end{align}
where $\sigma_t$ is the added diffusion noise and $\rv z_t$ is the noise-corrupted version of the  state $\rv x_k$, and $D_\theta$ is a learned conditional denoiser.  Thus, rather than directly learning a high-dimensional probability gradient, STORM learns to denoise the clean Earth system state from its noisy representation. This denoiser provides the learned conditional prior score required by the posterior sampling process in Eq.~\eqref{reverse:SDE}.

\begin{figure*}[h!]
    \centering
    \begin{subfigure}{.5\linewidth}
    \includegraphics[width=\linewidth]{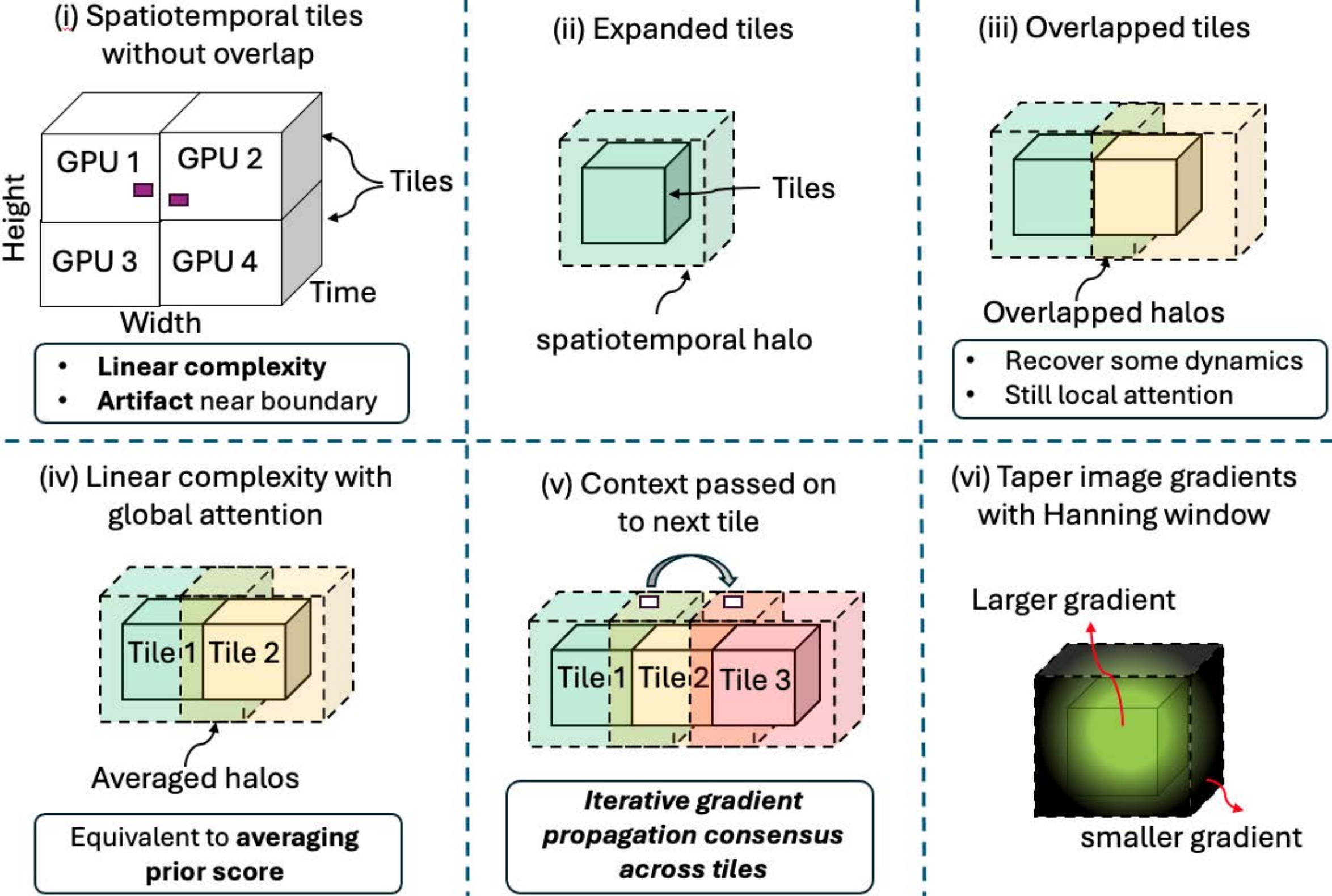}
    \caption{Overview of STORM Scaling Algorithm.
    }
    \end{subfigure}
    \quad\quad\quad
    \begin{subfigure}{.23\linewidth}
    \includegraphics[width=\linewidth]{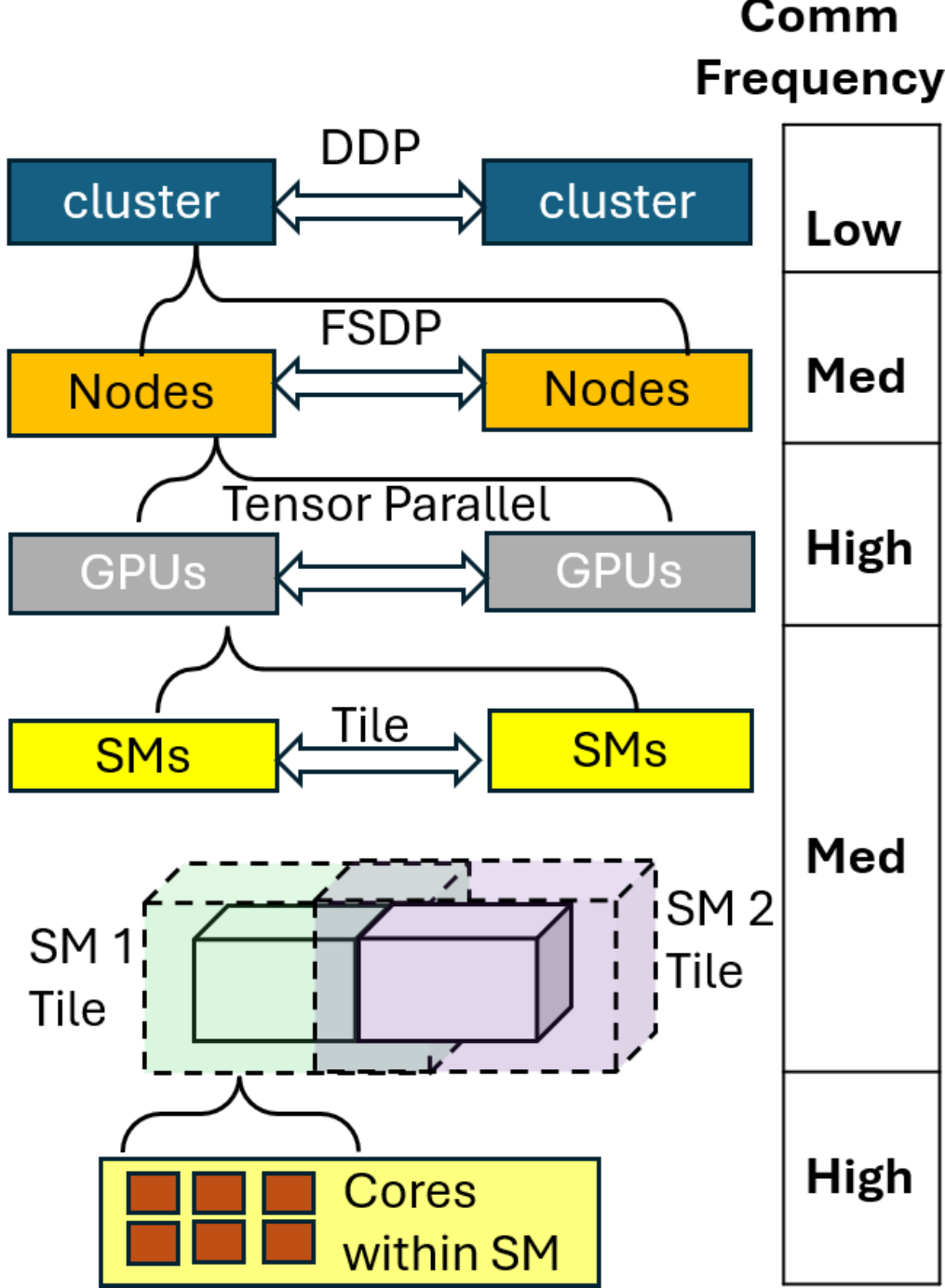}
    \caption{Hierarchical parallelism
    }
    \end{subfigure}    
\caption{\em (a) Tiling achieves linear complexity but limits interactions to local regions, while halo overlap improves continuity but remains local. STORM-GPA averages denoised outputs (gradients) in overlapping regions and propagates them iteratively across tiles, enabling global context with linear complexity. Hanning weighting stabilizes boundary interactions. (b) Hierarchical mapping of parallelism strategies to supercomputer hardware, with communication frequencies shown on the right.}
\label{fig:STORM-scaling}
\vspace{-0.5cm}
\end{figure*}

We parameterize this denoiser $D_\theta$ with the STORM-DiT architecture, a scalable spatiotemporal diffusion ViT architecture for high-resolution Earth system. The denoiser is trained to reconstruct the clean current state $\rv x_k$ from its noisy state $\rv z_t$ conditioned on the historical states:
\begin{equation}
\mathcal{L}(\theta) = \mathbb{E}\left[\| D_\theta(\rv z_t, \sigma_t ; \rv x_{k-K:k-1}) - \rv x_k \|^2 \right] \ ,
\end{equation}
and the denoiser in turn provides the prior score used by the one-stage posterior sampling process.

A central challenge is that the denoiser must simultaneously represent
high-resolution spatial structure and long temporal context. State-of-the-art
spatiotemporal ViTs such as TimeSformer have complexity
$\mathcal{O}(K^2N+KN^2)$, where $K$ is the temporal context length and
$N$ is the number of spatial tokens per temporal frame. The long temporal
context therefore multiplies the cost of global spatial attention.
STORM-DiT decouples space and time by compressing historical information
into a compact temporal representation, removing this multiplicative
coupling and reducing per-layer complexity to $\mathcal{O}(N^2)$ while
preserving global correlations. Table~\ref{table:scaling_comparison}
shows that STORM-DiT handles 28\% more spatiotemporal token scaling and
50\% lower runtime than TimeSformer.
% \vspace{-0.1cm}

Figure~\ref{fig:workflow-architecture}(b) summarizes its architecture. STORM-DiT first encodes both the noisy current state $z_t$ and clean historical states $\rv x_{k-K:k-1}$ into token embeddings for each variable independently, where each variable represents a different weather and climate {variable} such as temperature, humidity, and wind at each pressure level. Historical states are aggregated into a global representation using cross-attention first across the variable dimension and then across the temporal dimension (highlighted in red), while the noisy current state $z_t$ is aggregated only in the variable dimension, leveraging cross-attention's dual role as both correlation modeling and weighted aggregation. Additional context from calendar time, spatial \& temporal resolution and token spatial position are added to embedding for additional training context. Then at each transformer layer, STORM-DiT performs two complementary operations: (i) spatial self-attention on the current state, and (ii) cross-attention between the current state and the compressed temporal representation, where both operations have a complexity of $\mathcal{O}(N^2)$, leading to a global $\mathcal{O}(N^2)$ complexity.
% \vspace{-0.1cm}

To account for diffusion uncertainty, we introduce a {\em noise-gated attention mechanism} that modulates the relative contribution of spatial self- and temporal cross-attention using diffusion noise $\sigma_t$, and is highlighted by a red box in the figure. At high noise levels, STORM-DiT relies more on historical input and temporal cross-attention, while at low noise levels, it emphasizes spatial refinement of the current state and spatial self-attention. This design improves stability and accuracy during sampling.
Finally, model outputs are decoded into a denoised estimate of the {Earth system} state.

\subsection{STORM-GPA: Linear-Complexity Global Attention}
\label{sec:storm-linear-scaling}
STORM-DiT architecture removes the space--time coupling of conventional spatiotemporal attention, reducing complexity to $\mathcal{O}(N^2)$. However, its full-resolution global spatial attention remains quadratic and becomes prohibitive at extreme resolutions. We therefore introduce \emph{STORM Global Propagation Attention} (STORM-GPA), a linear-complexity scaling algorithm that achieves global context through iterative local gradient propagation across spatial tiles. Crucially, this propagation is integrated into the existing diffusion denoising steps and \textbf{introduces no additional iterations}.

A natural approach to achieve linear complexity is to partition the spatiotemporal volume into tiles and process each tile independently on separate GPUs~\cite{2025orbit2}, as illustrated in Figure~\ref{fig:STORM-scaling}(a, i). This reduces computation to linear complexity as long as the number of tiles scales linearly with resolution. However, this approach restricts attention to local tiles, leading to boundary artifacts due to missing cross-tile correlations. 
We extend each tile with spatiotemporal halos that overlap with neighboring tiles (Figure~\ref{fig:STORM-scaling}(a, ii-iii)). This improves local consistency but attention remains fundamentally local and global correlations are still missing.

\textbf{Global context through gradient propagation.}
The key insight behind STORM-GPA is that diffusion denoising provides gradient information for posterior inference. From Eq.~(\ref{eqn:denoise}), the denoising operator estimates the prior score, i.e., the gradient of the log-prior density. Thus, each denoising step performs a gradient update, and independently denoising each tile produces local gradient estimates.
To enforce consistency across tiles, we average denoised outputs in overlapping halos (Figure~\ref{fig:STORM-scaling}(a, iv)). Since denoising corresponds to estimating the prior score, which is a gradient operation, averaging denoised outputs is equivalent to averaging gradient estimates at shared boundaries. This enforces consistency in the overlap regions, communicates local context across neighboring tiles while maintaining linear complexity.

This shared context information is then propagated within each tile through attention: tokens in the interior attend to halo regions, allowing context received from neighboring tiles to influence the entire tile. As a result, each tile's update incorporates partial global context obtained from its neighbors.
Crucially, this propagation occurs naturally across the existing denoising steps, without introducing additional iterations (Figure~\ref{fig:STORM-scaling}(a, v)). At each denoising step, context propagates from tile to tile, progressively expanding its receptive field. Over successive denoising steps, this repeated local gradient propagation through averaging the halo regions and attention algorithm enables context from each tile to traverse the entire domain, effectively attending globally without explicit all-to-all communication or additional iterative procedures. As gradients across tiles converge, updates stabilize and propagation naturally diminishes. In this process, computations remain confined to tiles and their neighbors, yielding linear complexity.
To further improve stability and accelerate convergence, we apply a Hanning window (Figure~\ref{fig:STORM-scaling}(a, vi)) to weight gradient contributions, assigning higher confidence to central regions and lower weight near boundaries.

\subsection{Hierarchical Parallelism and System Scaling}
\label{sec:parallelism}
Achieving exascale performance requires not only scalable algorithms but also
a communication-aware mapping of algorithmic parallelism onto the hardware hierarchy. STORM therefore employs five levels of hierarchical parallelism, explicitly mapped to the five levels of Frontier's hardware hierarchy: sub-clusters, nodes, GPUs, streaming multiprocessors (SMs), and cores, as illustrated in Fig.~\ref{fig:STORM-scaling}(b). The guiding
principle is to match communication frequency and volume to hardware
bandwidth and latency, placing frequent communication on lower-latency
hardware and infrequent communication on higher-latency hardware.

\textbf{Level 1: Data parallelism (DDP) across sub-clusters.}
DDP shards training data across sub-clusters and requires relatively
infrequent communication. We therefore map DDP across sub-clusters, where the higher communication latency is amortized by the low synchronization
frequency.
\textbf{Level 2: Fully sharded data parallelism (FSDP) across nodes.}
FSDP partitions model parameters, optimizer states, and training data across nodes within each sub-cluster. Its moderate synchronization frequency is matched to the relatively high-bandwidth inter-node network. 
\textbf{Level 3: Tensor parallelism across GPUs.}
Tensor parallelism partitions model parameters across GPUs within a node. Because tensor-parallel operations require frequent synchronization, they are mapped within nodes to exploit high-bandwidth, low-latency GPU
interconnects.
\textbf{Level 4: STORM tile parallelism across SMs.}
STORM-GPA partitions the spatiotemporal domain into overlapping tiles and propagates context through attention and gradient averaging at the overlap regions at each diffusion step. We map these tiles within a GPU across its SMs rather than
across GPUs. This mapping is critical: distributing tiles across GPUs would turn frequent tile-level communication into costly inter-GPU communication and synchronization. Keeping tiles within a GPU allows neighboring tiles to exchange context through fast device memory and on-chip caching, preserving the locality of STORM-GPA while minimizing data movement.
\textbf{Level 5: Fine-grained attention parallelism across cores.}
Within each SM, fine-grained attention operations are distributed across cores and optimized using FlashAttention~\cite{dao2023flashattention2}.
These operations exploit registers and on-chip cache to reduce memory traffic and maximize arithmetic throughput.

Together, these parallelisms form a communication-aware hierarchy in which high-frequency communication remains close to computation, and lower-frequency synchronization is  pushed to higher hardware levels. This hardware--algorithm co-design enables STORM to minimize communication as the system scales to 74,400 GPUs while maintaining strong scaling efficiency at 96--99\% (see Fig.~\ref{fig:strong_scaling}).

To ensure that reported scaling reflects steady-state execution, STORM performs communication, kernel-cache, training-step, and data-staging warmup prior to timed runs, reducing initialization overheads and run-to-run variability, thereby improving the robustness and reproducibility of whole-application performance at exascale scale.
% To ensure that reported scaling reflects steady-state execution, STORM performs a sequence of readiness and warmup operations prior to timed runs, including communication initialization, persistent kernel-cache population, representative training-step execution, and local NVMe data staging. These steps remove one-time initialization costs from the critical training path and reduce run-to-run variability arising from communication setup, runtime initialization, and storage-system effects at exascale scale. Combined with continued evolution of the Frontier software environment, they contribute to robust and reproducible whole-application performance measurements.
For further efficiency, we incorporate system-level optimizations, including BF16 mixed precision, activation checkpointing and layer wrapping for improved memory footprint and throughput. In addition, we adopt hybrid-OP orthogonal parallelism from ORBIT~\cite{wang2024orbitoakridgebase}, alternating matrix row and column sharding to reduce peak memory use and communication.

\section{How Performance Was Measured}
\label{sec:performance-setup}

\textbf{Evaluation Overview.}
We evaluate STORM along two complementary dimensions:
\emph{computational scalability} and \emph{scientific capability}.
Computational experiments use the global ERA5 hourly dataset
with $2\times2$ spatial tokenization, enabling controlled evaluation of model,
spatial, temporal, and GPU scaling. Scientific experiments evaluate three
workloads: (1) hurricane prediction using ERA5 hourly data, (2) temperature
estimation using HRRR hourly data, and (3) long-term climate reanalysis
using weekly aggregated ERA5 data. 

\textbf{Model Configuration.}
We evaluate four model scales: a 10M-parameter model (embedding dimension 256, 6 layers), a 200M-parameter model (embedding dimension 1024, 8 layers), a 1B-parameter model (embedding dimension 2048, 8 layers),
and a 10B-parameter model (embedding dimension 3072, 40 layers).
\begin{table}[h!]
\centering
\footnotesize
\label{tab:scaling}
\begin{tabular}{cccc}
\toprule
\textbf{Method} & \textbf{Tokens} & \textbf{Memory (GB)} & \textbf{Time (s)} \\
\midrule
\multirow{3}{3cm}{Conventional ViT $(\mathcal{O}(K^2N^2))$} 
& 61K   & 2.0  & 0.007 \\
& 384K  & 12.0 & 0.225 \\
& 1.5M  & 46.0 & 3.560 \\
\midrule

\multirow{4}{3cm}{TimeSformer \\ $(\mathcal{O}(K^2N+N^2K))$} 
& 61K   & 1.4  & 0.003 \\
& 384K  & 7.6  & 0.016 \\
& 1.5M  & 29.0 & 0.058 \\
& 3.1M  & 55.0 & 0.080 \\
\midrule

\multirow{5}{3cm}{STORM-DiT ($\mathcal{O}(N^2)$)} 
& 61K   & 1.2  & 0.002 \\
& 384K  & 6.6  & 0.018 \\
& 1.5M  & 23.0 & 0.032 \\
& 3.1M  & 45.0 & 0.048 \\
& 3.8M  & 57.0 & 0.056 \\
\midrule

\multirow{6}{3cm}{STORM-DiT + GPA ($\mathcal{O}(N)$)} 
& 61K     & 0.5  & 0.006 \\
& 384K    & 1.8  & 0.010 \\
& 3.1M    & 3.1  & 0.055 \\
& 12.3M   & 3.1  & 0.220 \\
& 49.2M   & 11.6 & 0.520 \\
& \textbf{196.6M} & \textbf{46.0} & \textbf{1.540} \\
\bottomrule
\end{tabular}
\caption{\em Scaling comparison on a 10M-parameter model using 16 GPUs. STORM-DiT reduces space--time attention complexity to quadratic in spatial tokens, while STORM-GPA further reduces global attention to linear complexity, enabling two orders of magnitude longer token sequences.}
\label{table:scaling_comparison}
\vspace{-0.1cm}
\end{table}

\textbf{System Details.}
All experiments use Frontier supercomputer at Oak Ridge National Laboratory. Each node contains one 64-core AMD EPYC CPU and eight 64-GB GPUs, organized as four MI250X accelerators with each containing two GPU dies. Intra-node GPUs are connected by 50~GB/s Infinity Fabric, and nodes are interconnected by 100~GB/s Slingshot-11. The software stack uses PyTorch 2.10, ROCm 7.13 and
libfabric 2.3.1.

\begin{table*}[t]
\centering
\footnotesize
\begin{minipage}[t]{0.49\textwidth}
\centering
\textbf{(a) Temporal Scaling (Spatial size 32$\times$64)}
\vspace{2mm}

\begin{tabular}{ccccc}
\toprule
\textbf{Temporal} & \textbf{GPUs} & \textbf{Tiles} & \textbf{Model} & \textbf{Tokens} \\
\midrule
5K     & 16  & 1   & 10M  & 18M \\
17.5K  & 16  & 4   & 10M  & 63M \\
62.5K  & 16  & 16  & 10M  & 224M \\
127.5K & 64  & 16  & 10M  & 456M \\
177.5K & 128 & 16  & 10M  & 635M \\
45K    & 128 & 16  & 200M & 161M \\
15K    & 128 & 16  & 10B  & 54M \\
\bottomrule
\end{tabular}
\end{minipage}
\hfill
\begin{minipage}[t]{0.49\textwidth}
\centering
\textbf{(b) Spatial Scaling (1 Temporal Frame)}
\vspace{2mm}

\begin{tabular}{ccccc}
\toprule
\textbf{Spatial Size} & \textbf{GPUs} & \textbf{Tiles} & \textbf{Model} & \textbf{Tokens} \\
\midrule
1600$\times$3200   & 16  & 1   & 10M  & 10M \\
4800$\times$9600   & 16  & 16  & 10M  & 92M \\
14400$\times$28800 & 16  & 256 & 10M  & 829M \\
24000$\times$48000 & 64  & 256 & 10M  & 2.3B \\
30400$\times$60800 & 128 & 256 & 10M  & 3.7B \\
22400$\times$44800 & 128 & 256 & 200M & 2.0B \\
8000$\times$16000  & 128 & 256 & 10B  & 256M \\
\bottomrule
\end{tabular}
\end{minipage}
\caption{\em STORM maximal sequence scaling. Left: temporal scaling at fixed spatial resolution. Right: spatial scaling with one temporal frame. $\texttt{Tiles}=1$ denotes STORM-DiT only, while $\texttt{Tiles}>1$ activates linear-complexity tiling.}
\label{tab:maximal-sequence}
\vspace{-0.5cm}
\end{table*}

\textbf{Datasets.}
We evaluate STORM using two complementary Earth atmosphere datasets spanning global and regional scales.

\emph{ERA5 (global reanalysis)}~\cite{era5}. ERA5 provides hourly global atmospheric data at $\sim$28-km resolution ($720\times1440$ grid per temporal frame). We use 1978--2017 for training with 350K training samples and 2018--2020 for validation and testing. Each sample contains six variables: orography, land--sea mask, latitude, 10-m $u,v$ wind components, and 2-m temperature, forming tensors of size $720\times1440\times6\times K$, where $K$ is the temporal context length. For long-horizon climate reanalysis, we additionally aggregate the hourly data into weekly averages using a daily sliding window. The weekly dataset uses the same 1978--2017 training period with 14K training samples and 2018--2020 for validation/testing.

{\em High-Resolution Rapid Refresh (HRRR) regional data}~\cite{dowell2022hrrr}.
HRRR provides 5 km resolution hourly data over the continental United States region ($512 \times 1024$ grid). 
We use 2021–2023 for training and 2024 for validation and testing, totaling 26k  training data points. Each data point includes four variables (orography, land–sea mask, latitude, and 2-m temperature), forming tensors of size $512 \times 1024 \times 4 \times K$.

\textbf{Computing Performance Metrics.} We use the following performance metrics to evaluate our computing results:
\begin{itemize}[leftmargin=10pt]
\item \textbf{Memory footprint:} peak reserved GPU memory in gigabytes (GB) during training.
\item \textbf{Training and inference time-to-solution:}
average wall-clock time per data point during training and ensemble inference, averaged over multiple runs.
\item \textbf{Strong scaling efficiency:} speedup relative to a baseline of 64 nodes (512 GPUs), normalized by GPU count.
\item \textbf{Sustained throughput:} average achieved floating-point performance (FLOP/s) during training. The performance includes the whole application with IO. Reported in \mbox{ExaFLOPS} and measured by DeepSpeed FlopsProfiler.
\end{itemize}
\vspace{0.1cm}

\textbf{Scientific Performance Metrics.}
We compare three cases: the forward simulation without DA correction, STORM posterior sampling, and the reference analysis. Accuracy of the ensemble mean is measured using root-mean-square-error (RMSE), while uncertainty is quantified by ensemble standard deviation (STD). We quantify DA impact by (i) reduction in RMSE and STD compared to the forward simulation, and (ii) uncertainty reduction via variance reduction and spread–skill alignment.
These metrics are evaluated using ERA5 10-m wind speed for hurricane tracking, HRRR 2-m temperature for regional state estimation, and ERA5 2-m temperature for climate reanalysis.

\section{Performance results}
\label{sec:expriments}
\subsection{Computing Performance and Scalability}

\noindent\textbf{Scaling comparison across attention mechanisms.}
Table~\ref{table:scaling_comparison} compares training scalability on a 10M-parameter model using 16 GPUs with ERA5 data and a spatial token size of $2\times 2$. The results show a progression from quadratic global attention to scalable global modeling: conventional ViT and TimeSformer remain quadratic, STORM-DiT removes the space--time coupling and extends the feasible scale to 3.8M tokens. However, its quadratic spatial attention remains limiting. Adding STORM-GPA changes the scaling from $\mathcal{O}(N^2)$ to $\mathcal{O}(N)$, enabling 196.6M tokens, over 50$\times$ beyond the largest STORM-DiT sequence. Memory remains nearly constant over the tiled regime, demonstrating that the dominant cost shifts from global attention to local tile computation.

\textbf{Temporal and spatial maximal sequence scaling.}
Table~\ref{tab:maximal-sequence}  shows the extreme sequence lengths enabled by STORM-DiT + STORM-GPA on ERA5 dataset. In temporal scaling, the spatial domain is fixed at $32\times64$ while the temporal context increases. STORM reaches 177.5K temporal frames and 635M tokens on a 10M-parameter model. In spatial scaling, using one temporal frame, STORM reaches $30{,}400\times60{,}800$ spatial points, corresponding to 3.7B tokens. The spatial experiments extrapolate beyond ERA5's native resolution to characterize computational limits. Even with a 10B-parameter model, STORM supports 54M tokens in temporal scaling and 256M tokens in spatial scaling.
Note that $\texttt{Tiles}=1$ represents STORM-DiT without STORM-GPA, whereas $\texttt{Tiles}>1$ activates the STORM-GPA linear-complexity tiled implementation. As STORM-DiT decouples spatial and temporal attention, the two dimensions can scale independently rather than being coupled via a global spatiotemporal attention matrix.
\begin{figure}[h!]
\vspace{-0.4cm}
\centering    \includegraphics[width=.75\linewidth]{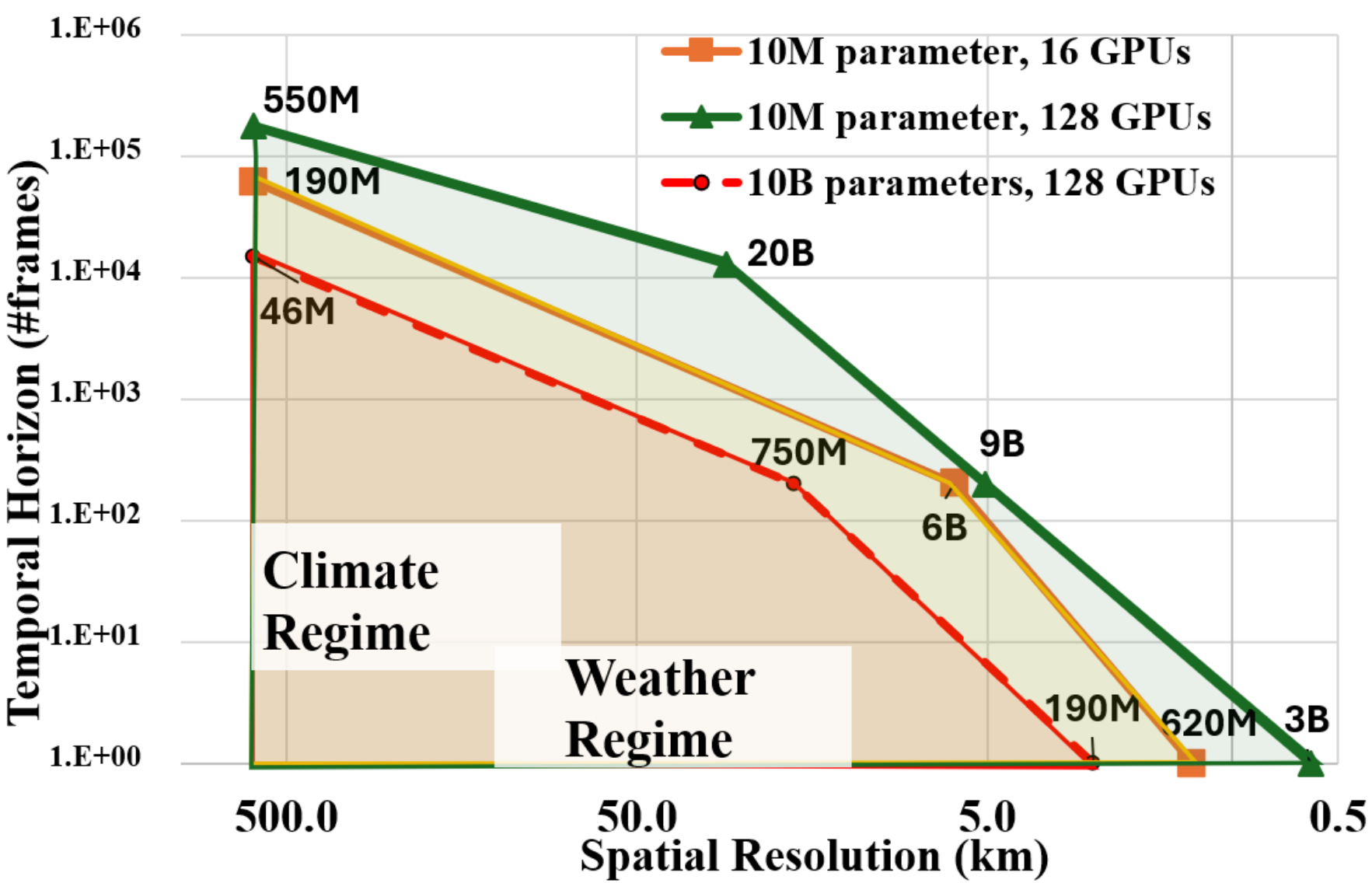}
\caption{\em Trade-off between spatial resolution and temporal context under fixed compute budgets. Each curve represents achievable spatiotemporal configurations for different model sizes and GPU counts. STORM expands the feasible frontier, enabling simultaneous high resolution (km scale) modeling and long temporal horizons with up to 20B tokens, bridging weather-scale forecasting and climate scale simulation.} 
\label{fig:weather_vs_climate}  
\vspace{-0.1cm}
\end{figure}

\textbf{Joint spatiotemporal scaling and new capability.}
Fig.~\ref{fig:weather_vs_climate} shows the achievable trade-off between spatial resolution and temporal context under fixed compute budgets. Conventional weather and climate models typically trade spatial resolution against temporal context: weather models operate at relatively high spatial resolution but over shorter time scales, while climate models target longer-term variability but generally operate at much coarser resolution. These regimes are fundamentally constrained by computational scaling limits.
STORM-DiT + STORM-GPA substantially expands this feasible region by independently scaling spatial and temporal dimensions, enabling regimes that combine high spatial resolution with long temporal context. For example, at 128 GPUs, the 10M-parameter model can achieve  $\sim$600 m global resolution or over 177K temporal frames, and critically, can jointly achieve meaningful regimes  such as 5 km resolution with 200 temporal frames.
This demonstrates a computational regime that is infeasible for conventional global-attention approaches.
\begin{figure}[h!]
\vspace{-0.3cm}
\centering    \includegraphics[width=.8\linewidth]
{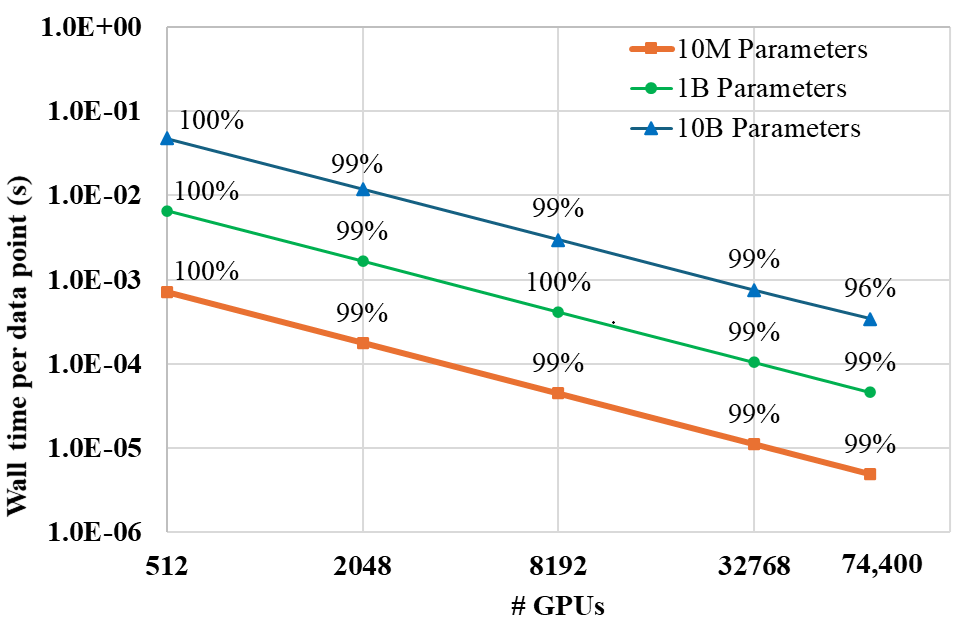}
\vspace{-0.1cm}
\caption{\em Strong scaling efficiencies across various model sizes, scaling to 74,400 GPUs with 96\% to 99\% strong scaling efficiencies with up to 6 exaFLOP sustained computing throughput at BF16 precision.}
\label{fig:strong_scaling} 
\vspace{-0.2cm}
\end{figure}

\textbf{Strong scaling performance.}
Fig.~\ref{fig:strong_scaling} shows strong scaling during training on ERA5 across 10M, 1B, and 10B models. STORM scales to 74,400 GPUs, the full Frontier supercomputer available during experiments, while maintaining 96--99\% strong-scaling efficiency and achieving up to 6.0 ExaFLOPS sustained BF16 throughput. This scalability results from the combined algorithmic and system-level co-design: STORM-GPA confines attention communication to local tiles and their neighbors, while the hierarchical parallelism maps communication patterns onto Frontier's hardware hierarchy.
\begin{table}[h!]
\centering
\footnotesize
\begin{tabular}{cccccc}
\toprule
\textbf{Model} & \textbf{GPUs} & \textbf{Ensemble Size} & \textbf{Wall Time (s)} \\
\midrule
10M  & 8     & 64      & 34.6  \\
10M  & 256   & 2048    & 34.5  \\
10M  & 4096  & 32768   & 34.1  \\
\midrule
10B  & 8     & 8       & 240.1  \\
10B  & 256   & 256     & 241.4  \\
10B  & 10000 & 10000   & 242.9  \\
\bottomrule
\end{tabular}
\caption{\em Large-scale ensemble generation performance on ERA5. Wall-clock time remains nearly constant as ensemble size increases, demonstrating efficient sampling across GPUs.}
\label{tab:ensemble}
\vspace{-0.3cm}
\end{table}
% \vspace{-0.3cm}

\textbf{Large-scale ensemble generation.}
Table~\ref{tab:ensemble} demonstrates STORM's capability for large-scale ensemble inference over 80 diffusion steps at both 10M and 10B-parameter model sizes. As the number of GPUs increases, the ensemble size scales proportionally while wall-clock time remains nearly constant. For the 10M model, ensemble size increases from 64 to over 32K samples with essentially unchanged runtime ($\sim$34 seconds). Similarly, the 10B model scales to 10K ensemble members in $\sim$243 seconds. These results demonstrate that STORM can increase ensemble size nearly proportionally with GPU resources without increasing time-to-solution, enabling large-ensemble UQ at practical latency.
\begin{table}[h!]
\centering
\begin{tabular}{cccc}
\toprule
Hurricane & w/o DA & DA with 10\% obs & DA with 20\% obs \\
\midrule
Michael & 1.91 & 1.21 & 0.76  \\
Teddy   & 1.89 & 1.38 & 0.84  \\
Laura   & 1.62 & 0.96 & 0.75  \\
Delta   & 2.13 & 2.12 & 0.90  \\
\bottomrule
\end{tabular}
\caption{\em
CRPS reported in degrees of latitude/longitude for hurricane track prediction and a comparison among no observational correction, 10\% and 20\% observations. Lower is better. STORM-OneStage consistently reduces probabilistic track error as more observations are assimilated, with up to $\sim$60\% improvement using only 20\% of observations.}
\label{tab:hurricane_crps}
\vspace{-0.5cm}
\end{table}

\subsection{Science Results}

STORM enables two complementary scientific capabilities that cannot be achieved with conventional DA: large-scale, low-latency probabilistic inference through massively parallel posterior
sampling, and long-context state inference through direct conditioning
on extended historical states. The first enables large ensembles for improved UQ especially for extreme events, while the second provides long temporal context for capturing slowly evolving Earth-system dynamics and improves
long-term climate reanalysis.

We evaluate these capabilities across three scientific applications: global hurricane prediction, regional high-resolution temperature estimation, and long-term climate reanalysis. The first two test whether posterior sampling at scale improves accuracy and UQ under sparse observations, while the third tests whether extended temporal context enabled by our one-stage formulation provides measurable benefits for climate prediction and reconstruction.
Additional science results that are not covered in the paper due to page limits are available at \url{https://sites.google.com/view/exagenai}.

\begin{figure*}[t]
    \centering  
    \begin{minipage}[c]{0.69\textwidth}
    \centering   \includegraphics[width=\linewidth]{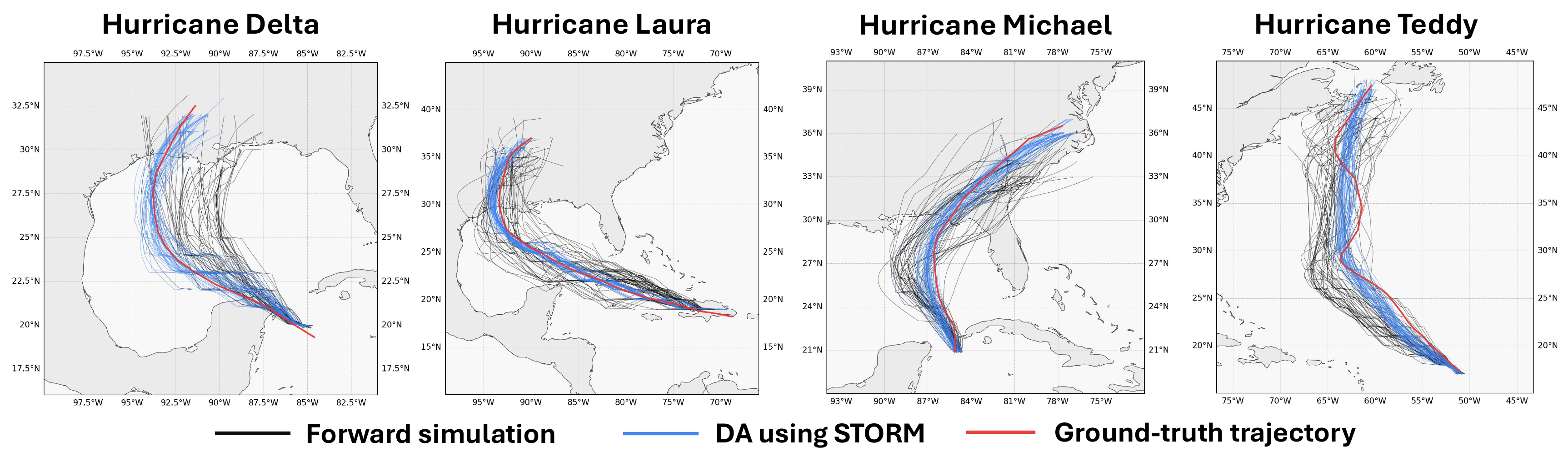}
    \end{minipage}
    \hfill
\begin{minipage}[c]{0.3\textwidth}
    \caption{\em \textbf{Hurricane track skill}: Ensemble trajectories for four hurricanes. Black: Forward simulation ensembles; blue: STORM-OneStage posterior ensembles using 20\% observations; red: ground truth reference published in \cite{nws_hurricane_kmz_set}. STORM reduces ensemble spread and improves agreement with reference trajectories.}
    \label{fig:era5_path}
\end{minipage}
\vspace{-0.3cm}
\end{figure*}

\textbf{{Global DA for hurricane tracking.}}
We evaluate STORM on four hurricanes (Michael'18, Laura'20, Delta'20, and Teddy'20), including multiple rapid-intensification events that are difficult to capture with current forecasting systems. While forecasting hurricane tracks are generally well constrained by environmental steering flows, accurately estimating intensity remains challenging because it depends on multiscale interactions between large-scale circulation and small-scale inner-core dynamics that are poorly resolved and weakly constrained by sparse observations.

\begin{figure*}[t]
    \centering    
\includegraphics[width=0.65\linewidth]{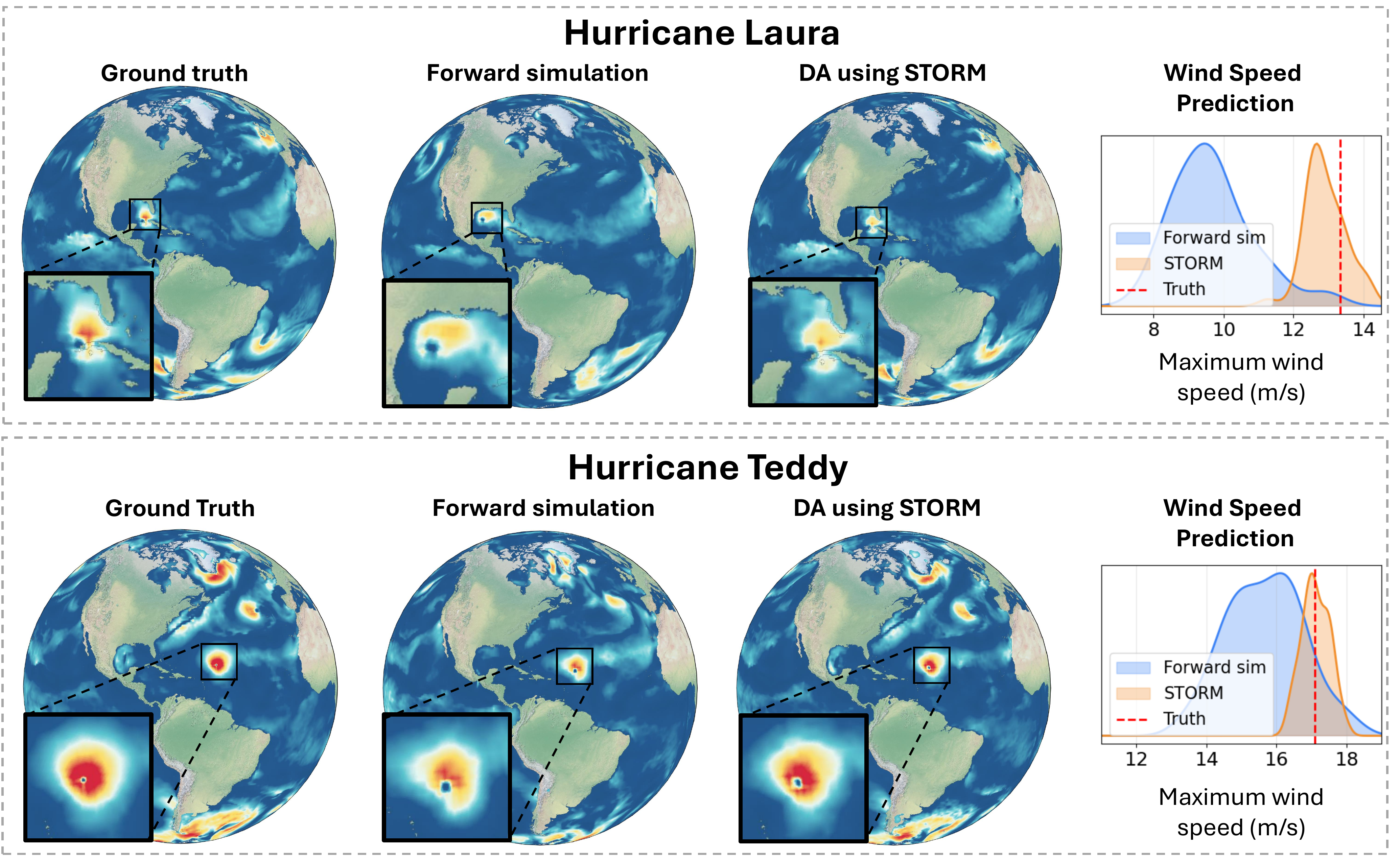}
    \caption{\em \textbf{Hurricane intensity skill}: Comparison of forward simulation and STORM-OneStage posterior ensembles for Laura and Teddy hurricanes. The last column in each row displays kernel density estimates (KDEs) of the maximum wind speeds computed across the ensemble, with the blue curve corresponding to forward simulation ensemble, and orange curve for STORM-OneStage posterior ensemble. The dashed red line represents ground truth hurricane intensity. Comparisons are aided by figure insets that zoom into each hurricane and additional results are provided on \url{https://sites.google.com/view/exagenai}. Note that STORM-OneStage Posterior sampling shifts the ensemble toward the observed intensity, recovering extreme wind speeds underestimated by forward simulation forecast.}
    \label{fig:era5_snapshot}
    \vspace{-0.5cm}
\end{figure*}

We distinguish two inference settings. Forward simulation results are sampled from the learned prior distribution, $S_{\text{prior}}$ in Eq.~\eqref{score_decomp}, without observational correction. STORM-OneStage results are sampled from the posterior distribution, $S_{\text{posterior}}$, where the observation likelihood corrects the learned prior. This allows us to directly measure the scientific benefit of observational conditioning.

Table~\ref{tab:hurricane_crps} quantifies track inference using the Continuous Ranked Probability Score (CRPS). Across all four hurricanes, STORM posterior results consistently improve tracking and reduce ensemble uncertainty. With only 20\% of observations, CRPS improves by up to $\sim$60\% relative to the forward simulation result. Substantial improvements remain even with only 10\% observations, demonstrating that STORM extracts useful state information from sparse observations while maintaining  probabilistic posterior inference.

Figure~\ref{fig:era5_path} further illustrates these improvements. The forward simulation ensembles exhibit large spread and occasional track errors, whereas STORM-OneStage posterior samples exhibit reduced spread and concentrate around the reference trajectories from \cite{nws_hurricane_kmz_set}. 
Thus, STORM-OneStage improves both track accuracy and the concentration of posterior uncertainty around the observed trajectory.

More importantly, STORM substantially improves hurricane intensity, a persistent challenge for forecasting systems. Fig.~\ref{fig:era5_snapshot} shows that forward simulation ensembles systematically underestimate peak wind speeds, particularly during rapid intensification. For example, Hurricane Laura's peak intensity is underestimated by approximately 4~m/s. In contrast, STORM-OneStage DA shifts both the ensemble mean and spread toward the ERA5 reference, substantially improving the representation of peak intensity. The qualitative comparison further show reduced along-track and structural errors, indicating that observational conditioning corrects both the location and strength of the simulated hurricane. These results demonstrate STORM can use partial observations to recover extreme hurricane states that are poorly represented by the forecast, rather than merely reducing prediction uncertainty.

\begin{figure*}[h!]
    \centering
\includegraphics[width=0.8\linewidth]{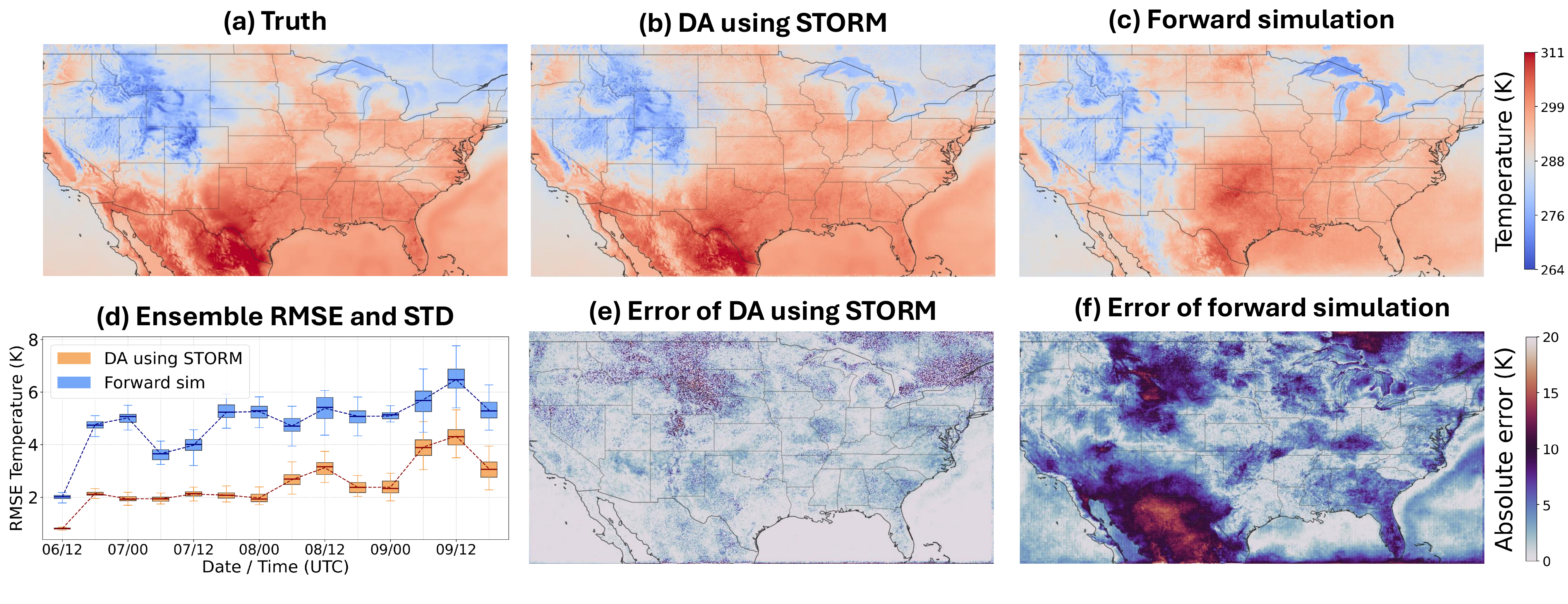}
\vspace{-0.3cm}
\caption{\em \textbf{Regional high-resolution temperature estimation}: 
(a) Ground truth 2-m temperature at 72-hour lead time. 
(b) STORM-OneStage posterior mean with 50\% observations. 
(c) Forward simulation ensemble mean. 
(d) RMSE evolution with ensemble spread, showing error reduction and improved UQ with STORM-OneStage.
(e,f) Corresponding absolute errors. }
    \label{fig:hrrr_snapshot}
    \vspace{-0.4cm}
\end{figure*}

\textbf{{Regional DA for high-resolution temperature estimation.}} 
We next evaluate STORM on 2-m temperature prediction at 5-km resolution over the continental United States. At this resolution, accurate state estimation requires resolving strong spatial heterogeneity and fast-evolving processes driven by terrain, land–atmosphere coupling, and diurnal forcing.
Conventional high-resolution DA is constrained by the cost of generating large forecast ensembles, limiting ensemble size and the representation of uncertainty and nonlinear error growth. STORM addresses this limitation through large-ensemble posterior sampling for high-resolution state inference.

We fine-tune STORM on HRRR data for 2021-2023 and evaluate both forward simulation ($S_{\text{prior}}$) and STORM posterior sampling initialized on 6 May 2024 at 00 UTC. The STORM posterior sampling assimilates synthetic 2-m temperature observations over 50\% of the grid, corresponding to an effective observational resolution of $\sim$7 km, significantly denser than typical surface networks (20–30 km spacing).

Figure~\ref{fig:hrrr_snapshot} shows results at a 72-hour lead time. Forward simulation exhibit substantial regional errors, particularly in areas with strong diurnal forcing. For example, diurnal heating along the US–Mexico border is underestimated by more than 10~K, reflecting the difficulty of capturing localized temperature extremes from forward simulation alone.
In contrast, STORM posterior sampling effectively corrects these errors through observational conditioning. The posterior temperature field closely matches the HRRR reference, with large regional biases largely eliminated. Panel (d) quantifies error evolution and uncertainty across time, showing that STORM consistently reduces RMSE by approximately 2–3~K relative to the forward simulation baseline.

\textbf{Climate reanalysis and reconstruction of extreme events.} DA is central not only to weather forecast but also to climate science. Historical observations are spatially incomplete and heterogeneous; DA combines these observations with a dynamical prior to produce reanalysis, a spatially complete reconstruction of the evolving Earth system. Reanalysis supports the study of atmospheric dynamics, long-term variability and trends, extreme events, and the evaluation and initialization of climate models \cite{buizza2018advancing, era5, gelaro2017merra2}. We therefore evaluate STORM in a reanalysis-style benchmark for global 2-m temperature at $1.0^\circ$ resolution. STORM is trained on ERA5 from 1978--2017 and evaluated exclusively on the held-out 2018--2020 period using weekly-averaged temperature.
\begin{figure}[h!]
\vspace{-0.1cm}
    \centering
\includegraphics[width=0.75\linewidth]{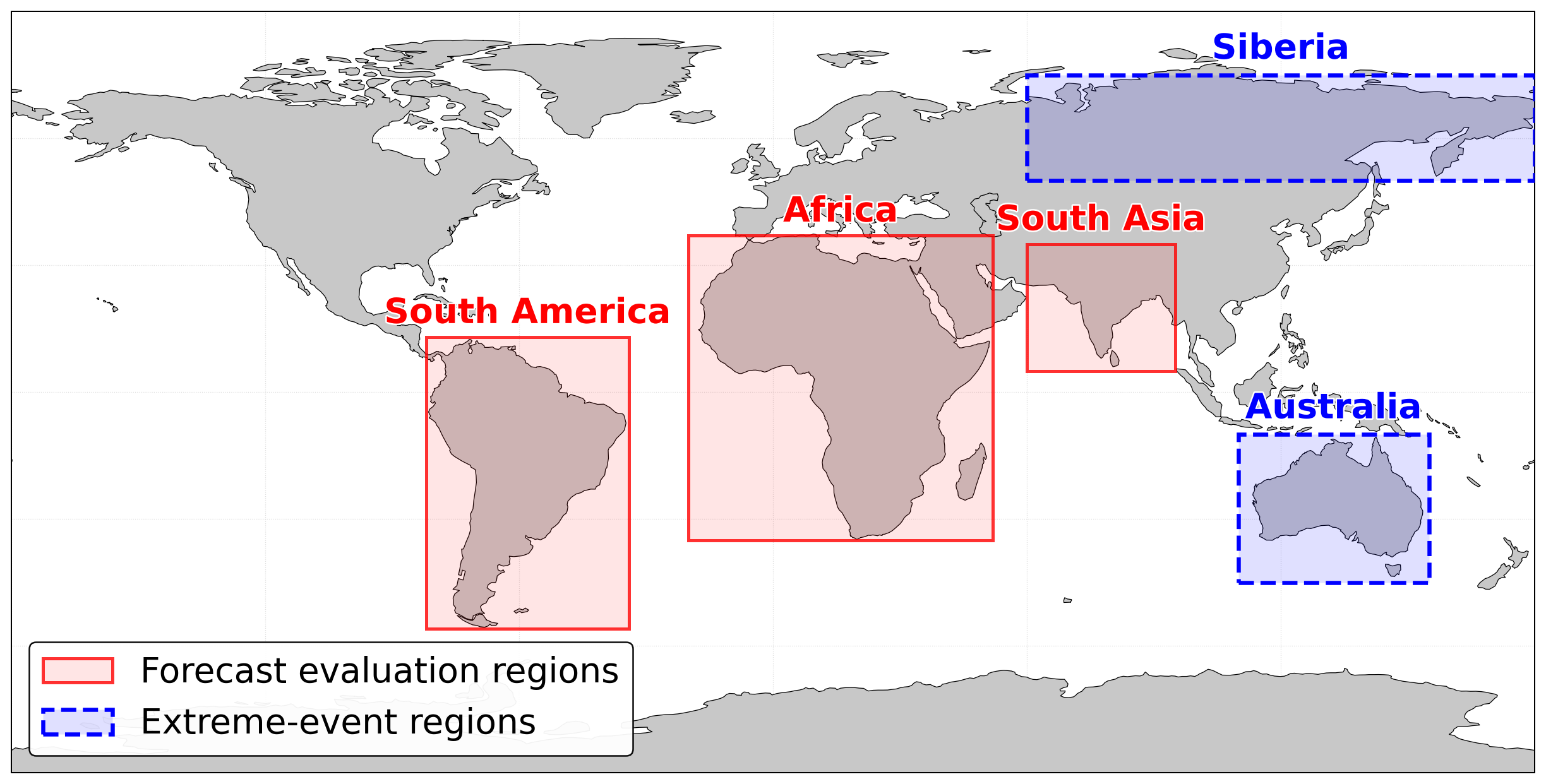}
    \caption{\em \textbf{Climate evaluation regions.} Regional RMSE is evaluated over Africa, South America, and South Asia (red boxes) in Figs.~\ref{fig:climate_noda}–-\ref{fig:climate_da}. Blue boxes denote Australia and Siberia used for the extreme-event evaluation in Fig.~\ref{fig:climate_event}.}
    \label{fig:world}
    \vspace{-0.5cm}
\end{figure}

\begin{figure}[h!]
    \centering
\includegraphics[width=0.95\linewidth]{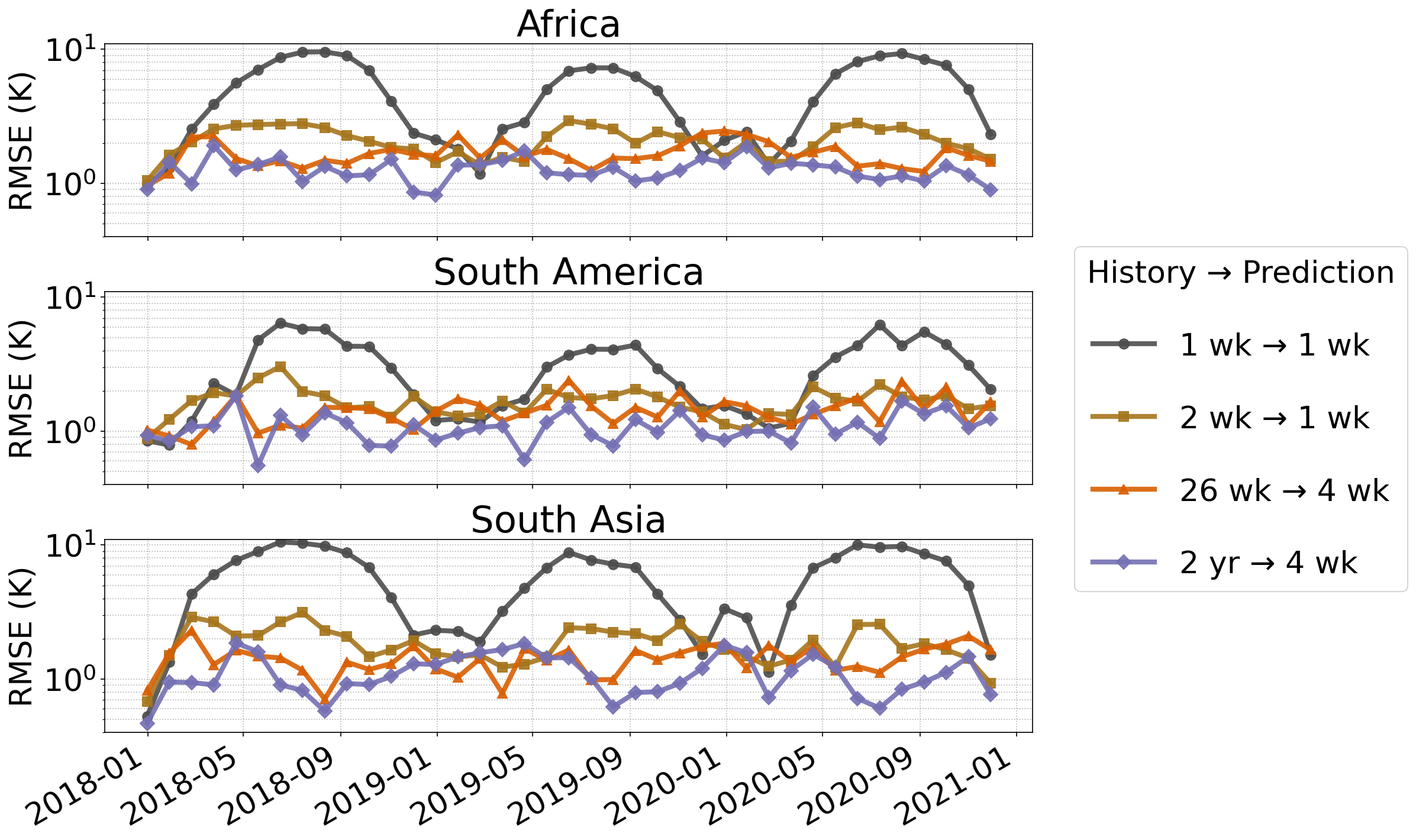}
    \caption{\em \textbf{Long temporal context improves STORM's forecast.} Weekly 2-m temperature RMSE over the held-out 2018–2020 testing period for four input/output configurations across Africa, South America, and South Asia. The 2yr-input-4wk-output model consistently achieves the lowest RMSE, demonstrating that multi-year temporal context improves the forecast used by STORM for posterior inference.}
    \label{fig:climate_noda}
    \vspace{-0.5cm}
\end{figure}

We first evaluate how temporal context affects the forward simulation, which provides the foundation for posterior inference. We compare four configurations with historical context lengths $K$ ranging from one week to two years. The first two models use 1 or 2 weeks of history to predict the following week, while the longer-context models use 26 weeks or 2 years of history to predict the following four weeks, which are then rolled out to simulate the full 2018--2020 period. Fig.~\ref{fig:climate_noda} shows that the two-year-context model consistently achieves the lowest RMSE across Africa, South America, and South Asia. These geographically separated regions, shown in Fig.~\ref{fig:world}, span distinct climate regimes, demonstrating that the benefit of long temporal context is not confined to a particular region. This result establishes a direct scientific benefit of STORM's long-context capability: extending temporal context from weeks to multiple years improves climate prediction accuracy rather than merely increasing computational scale. 
Unlike conventional sequential DA, which advances the forecast from the preceding state at each assimilation step, STORM can directly condition on extended historical states, enabling this long-context climate capability.

\begin{figure}[h!]
% \vspace{-0.3cm}
    \centering
\includegraphics[width=0.95\linewidth]{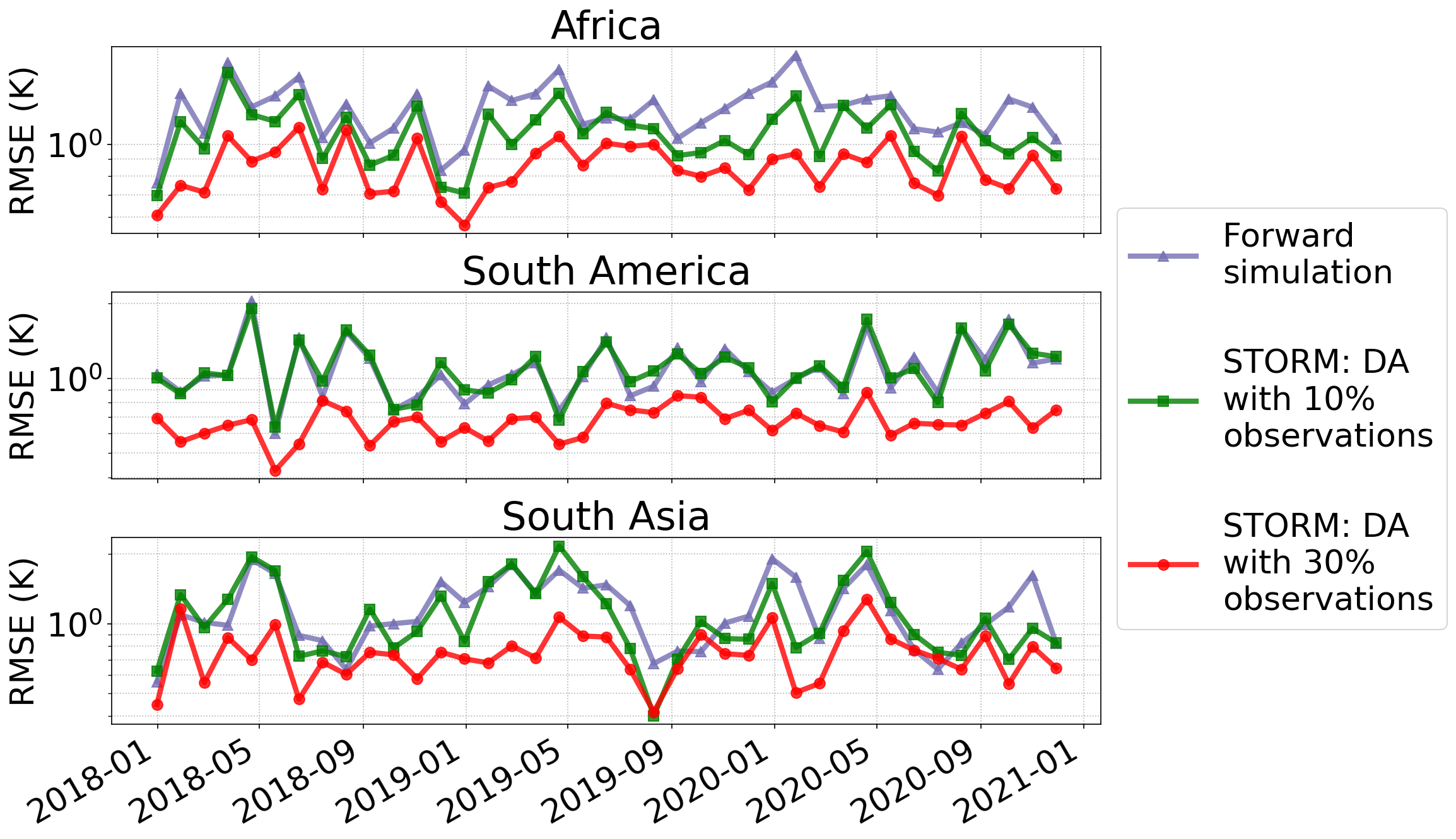}
    \caption{\em \textbf{Climate reanalysis improves with observational coverage.} Regional weekly 2-m temperature RMSE for the STORM model using two years of historical context to predict  the held-out 2018–2020 period. Black: forward simulation; colored curves: STORM posterior estimates using 10\% and 30\% of the observations. RMSE decreases systematically with increased observational coverage across all regions, demonstrating accurate reconstruction from incomplete observations.}
    \label{fig:climate_da}
    \vspace{-0.3cm}
\end{figure}

We next evaluate whether this improved long-context forecast can support accurate reanalysis from incomplete observations. Using the best performing
two-year-context model as the prior, we assimilate observations subsampled exclusively from the held-out ERA5 test period at observation fractions of 10\% and 30\%. Fig.~\ref{fig:climate_da} shows that regional RMSE systematically decreases as more observations are assimilated across all three regions. STORM therefore combines long temporal context with incomplete observations to reconstruct more accurate climate states, demonstrating the data--model synthesis capability required for long-term climate reanalysis.

\begin{figure}[h!]
\vspace{-0.3cm}
    \centering
\includegraphics[width=0.95\linewidth]{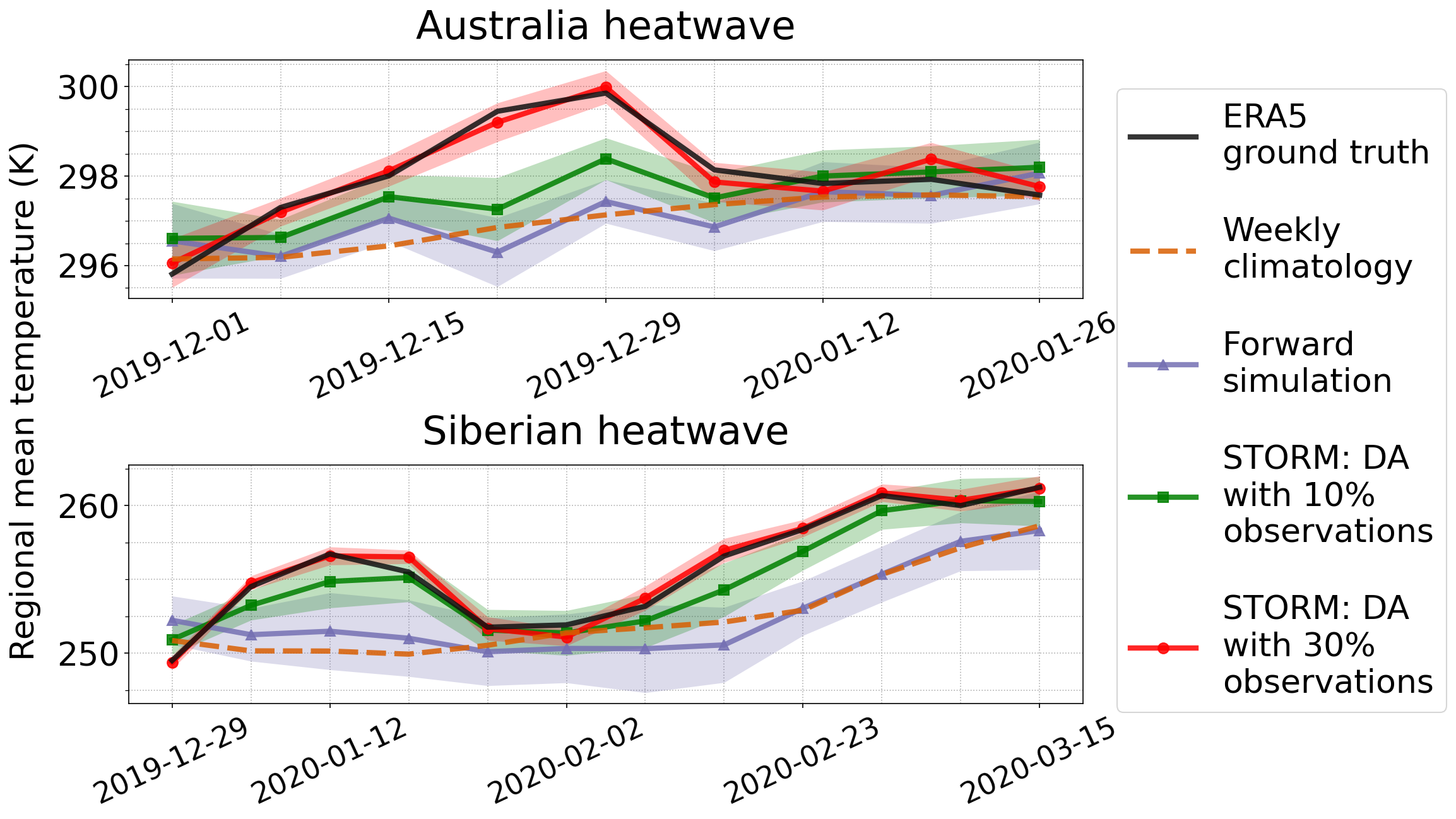}
    \caption{\em \textbf{Reconstruction of extreme temperature events with quantified uncertainties.} Regional weekly 2-m temperature during the 2019–2020 Australian Black Summer heatwave (top) \cite{BOM2020AustraliaSummer} and prolonged Siberian warmth in 2020 (bottom) \cite{WMO2020SiberianHeat}. The forward simulation largely follows climatological behavior and underestimates the extreme temperature evolution, whereas STORM progressively shifts the posterior toward the ERA5 reference as observational coverage increases. Shaded bands denote ensemble uncertainty, which narrows as observations constrain the inferred climate state.}
    \label{fig:climate_event}
    \vspace{-0.4cm}
\end{figure}

Finally, we examine two major temperature extremes within the held-out period: the 2019–2020 Australian Black Summer heatwave \cite{BOM2020AustraliaSummer} and the prolonged Siberian warmth of early 2020 \cite{WMO2020SiberianHeat}. These cases are particularly challenging because rare climate extremes can depart substantially from the climatological behavior learned by a forecast model. As shown in Fig.~\ref{fig:climate_event}, the forward simulation largely follows the climatological evolution and underestimates the observed extremes. Assimilating even partial observations shifts the posterior toward the ERA5 reference, while increasing observational coverage further improves the reconstruction and reduces posterior uncertainty. These results demonstrate that scalable probabilistic DA can recover extreme climate states missed by the forecast while simultaneously quantifying uncertainty in the reconstructed state.

\section{Implication}
\label{sec:implication}

\textbf{New computational regime enabled by STORM.} STORM expands the feasible computational regime for Earth system state inference by jointly supporting large ensembles, long temporal context, high spatial resolution, and low-latency inference. Conventional DA must trade among these capabilities because increasing resolution or ensemble size rapidly increases forward simulation and data storage, movement and I/O costs. By integrating a one-stage generative DA formulation with linear-complexity global attention, STORM removes key scaling barriers and enables these capabilities simultaneously at exascale.

\textbf{New climate science capabilities enabled by STORM.} The new computational regime from STORM enables scientific capabilities that are infeasible with conventional DA. STORM's large-scale posterior sampling enables tens of thousands of ensemble members for rapid probabilistic inference and UQ. In addition, STORM's one-stage DA formulation and scaling architecture and algorithm enables long temporal context, which provides information for slowly evolving Earth-system dynamics that is unavailable for conventional DA, which is conditioned only on a single preceding state. 
Our climate experiments demonstrate that this capability has measurable scientific value: increasing temporal context from weeks to two years improves long-term climate prediction, while combining the long-context prior with incomplete observations improves climate state reconstruction. STORM further demonstrates that observational conditioning can recover extreme states poorly represented by the forward simulation prior, including hurricane rapid intensification and regional temperature extremes. Together, these results establish scalable probabilistic state inference as a complementary exascale capability to forward Earth system simulation.

\textbf{Exascale generative AI as a scientific enabler.} STORM transforms data assimilation into a massively parallel generative inference workload that can efficiently utilize supercomputing systems. By scaling to 74,400 GPUs with 96--99\% strong scaling efficiency and tens of billions of spatiotemporal tokens and exceptionally long temporal context, this enables efficient processing of rapidly growing Earth system datasets (e.g., $\sim$700~GB/day from HRRR) and supports low latency assimilation at scale. 

Beyond climate modeling, the methodological and computational innovations are broadly applicable to applications requiring high-dimensional spatiotemporal data modeling and inference, including fusion energy, astrophysics, subsurface flow, and materials science, all facing similar challenges of high-dimensional state estimation under partial observations and complex dynamics. By enabling linear-complexity global attention and scalable generative posterior inference and communication-aware exascale execution STORM provides a framework for transforming traditionally memory- and communication-intensive workflows into compute-efficient and highly parallel workloads.

\section{Acknowledgment}

The authors gratefully acknowledge the AMD and Hewlett Packard Enterprise (HPE) teams for their contributions to extreme-scale execution of STORM on Frontier. In particular, we thank their efforts in diagnosing and mitigating large-scale runtime behavior associated with lazy/JIT initialization, communication-library startup, memory allocation, and execution stability, as well as for optimizing the runtime environment and data movement for reliable full-system experiments. These contributions were important in enabling stable and reproducible performance measurements at the scale of the full Frontier system.

This research was supported by the U.S. Department of Energy (DOE), Office of Science, Office of Advanced Scientific Computing Research (ASCR), Applied Mathematics Program under contracts ERKJ443 and ERKJ388, and through the SciDAC Institute LEADS. Additional support was provided by the Oak Ridge National Laboratory (ORNL) Artificial Intelligence Initiative, sponsored by the Laboratory Directed Research and Development (LDRD) Program.
This work was also supported by Dan Lu’s Early Career Project, funded by the DOE Office of Biological and Environmental Research (BER).
Feng Bao acknowledges support from the U.S. National Science Foundation (NSF) under grant DMS-2142672 and from the DOE Office of Science, ASCR Applied Mathematics Program under grant DE-SC0025412. Hristo Chipilski acknowledges support from Florida State University’s CRC Seed Grant (047080). Peter Jan van Leeuwen acknowledges support from the National Oceanic and Atmospheric Administration (NOAA) project CADRE (NA24OARX459C0002-T1-01) and the Department of Defense (DOD) project RAM-HORNS (N00014-24-2017).
This research used resources of the Oak Ridge Leadership Computing Facility at Oak Ridge National Laboratory, which is supported by the DOE Office of Science, ASCR, under Contract No. DE-AC05-00OR22725.

\bibliographystyle{IEEEtran}
\bibliography{main}

\end{document}